%% file: _main.tex
\input{_constants}
\arxiv 

\pdfoutput=1
\documentclass[10pt,twocolumn,letterpaper]{article}
\input{cvpr_header}
\begin{document}
\title{\paperTitle}
\author{\authorBlock}
\maketitle

\input{00_abstract}
\input{01_intro}
\input{02_related}

\input{03_method}
\input{04_experiments}

\input{10_conclusion}

{\small
\bibliographystyle{unsrt}
\bibliography{11_references}
}

\ifarxiv \clearpage \input{12_appendix} \fi

\end{document}

%% file: _constants.tex
\def\cvprPaperID{1519}
\def\confName{ICCV}
\def\confYear{2023}

\def\paperTitle{BEVBert: Multimodal Map Pre-training for Language-guided Navigation}

\def\authorBlock{
    Dong An$^{1,2}$ \ \
    Yuankai Qi$^{3}$ \ \
    Yangguang Li$^{4}$ \ \
    Yan Huang$^{1\dag}$ \ \
    Liang Wang$^{1}$ \ \
    Tieniu Tan$^{1,5}$ \ \
    Jing Shao$^{6\dag}$ \ \  \\
    $^{1}$Institute of Automation, Chinese Academy of Sciences \ \ 
    $^{2}$School of Future Technology, UCAS \\
    $^{3}$Australian Institute for Machine Learning, University of Adelaide \ \ 
    $^{4}$SenseTime Research \\
    $^{5}$Nanjing University \ \ 
    $^{6}$Shanghai AI Laboratory \\
    {\tt\small \url{https://github.com/MarSaKi/VLN-BEVBert}}
}

\newif\ifreview 
\newif\ifarxiv \newcommand{\arxiv}{\arxivtrue}
\newif\ifcamera 
\newif\ifrebuttal 

%% file: cvpr_header.tex
\ifreview \usepackage[review]{cvpr} \fi
\ifarxiv \usepackage[pagenumbers]{cvpr} \fi
\ifrebuttal \usepackage[rebuttal]{cvpr} \fi
\ifcamera \usepackage{cvpr} \fi

\usepackage{graphicx}
\usepackage{amsmath}
\usepackage{amssymb}
\usepackage{booktabs}

\input{_macros}  

\usepackage{xr-hyper}

\makeatletter
\newcommand*{\addFileDependency}[1]{
  \typeout{(#1)}
  \@addtofilelist{#1}
  \IfFileExists{#1}{}{\typeout{No file #1.}}
}

\makeatother

\usepackage[pagebackref,breaklinks,colorlinks]{hyperref}
\usepackage[capitalize]{cleveref}
\crefname{section}{Sec.}{Secs.}
\crefname{table}{Table}{Tables}
\crefname{figure}{Fig.}{Figs.}

%% file: _macros.tex
\usepackage{times}
\usepackage{microtype}
\usepackage{epsfig}
\usepackage[table,xcdraw]{xcolor}
\usepackage{caption}
\usepackage{float}
\usepackage{placeins}
\usepackage{color, colortbl}
\usepackage{stfloats}
\usepackage{enumitem}
\usepackage{tabularx}
\usepackage{xstring}
\usepackage{multirow,booktabs,makecell}
\usepackage{xspace}
\usepackage{url}
\usepackage{subcaption}
\usepackage{xcolor}
\usepackage[hang,flushmargin]{footmisc}
\usepackage{wrapfig}
\usepackage{adjustbox}
\usepackage{gensymb}

\ifcamera \usepackage[accsupp]{axessibility} \fi

% ...

\newcommand{\nbf}[1]{{\noindent \textbf{#1.}}}

\ifarxiv  \fi

\newcommand{\R}[1]{{%
    \textbf{%
        \ifstrequal{#1}{1}{\textcolor{red}{R#1}}{%
        \ifstrequal{#1}{2}{\textcolor{blue}{R#1}}{%
        \ifstrequal{#1}{3}{\textcolor{teal}{R#1}}{%
        \ifstrequal{#1}{4}{\textcolor{magenta}{R#1}}{%
                           \textcolor{cyan}{R#1}%
        }}}}%
    }%
}}

\usepackage{calc}
\newlength\myheight
\newlength\mydepth
\settototalheight\myheight{Xygp}
\newcommand*\inlinegraphics[1]{%
  \settototalheight\myheight{Xygp}%
  \settodepth\mydepth{Xygp}%
  \raisebox{-\mydepth}{\includegraphics[height=\myheight]{#1}}%
}

%% file: 00_abstract.tex
\begin{abstract}
Large-scale pre-training has shown promising results on the vision-and-language navigation (VLN) task.
However, most existing pre-training methods employ discrete panoramas to learn visual-textual associations. 
This requires the model to implicitly correlate incomplete, duplicate observations within the panoramas, which may impair an agent's spatial understanding. 
Thus, we propose a new map-based pre-training paradigm that is spatial-aware for use in VLN. 
Concretely, we build a local metric map to explicitly aggregate incomplete observations and remove duplicates, while modeling navigation dependency in a global topological map. 
This hybrid design can balance the demand of VLN for both short-term reasoning and long-term planning. 
Then, based on the hybrid map, we devise a pre-training framework to learn a multimodal map representation, which enhances spatial-aware cross-modal reasoning thereby facilitating the language-guided navigation goal.
Extensive experiments demonstrate the effectiveness of the map-based pre-training route for VLN, and the proposed method achieves state-of-the-art on four VLN benchmarks.
{\let\thefootnote\relax\footnote{$^{\dag}$Corresponding authors.}}
\end{abstract}

%% file: 01_intro.tex
\vspace{-3mm}
\section{Introduction}

\begin{figure}[!htbp]
\centering
\includegraphics[width=0.46\textwidth]{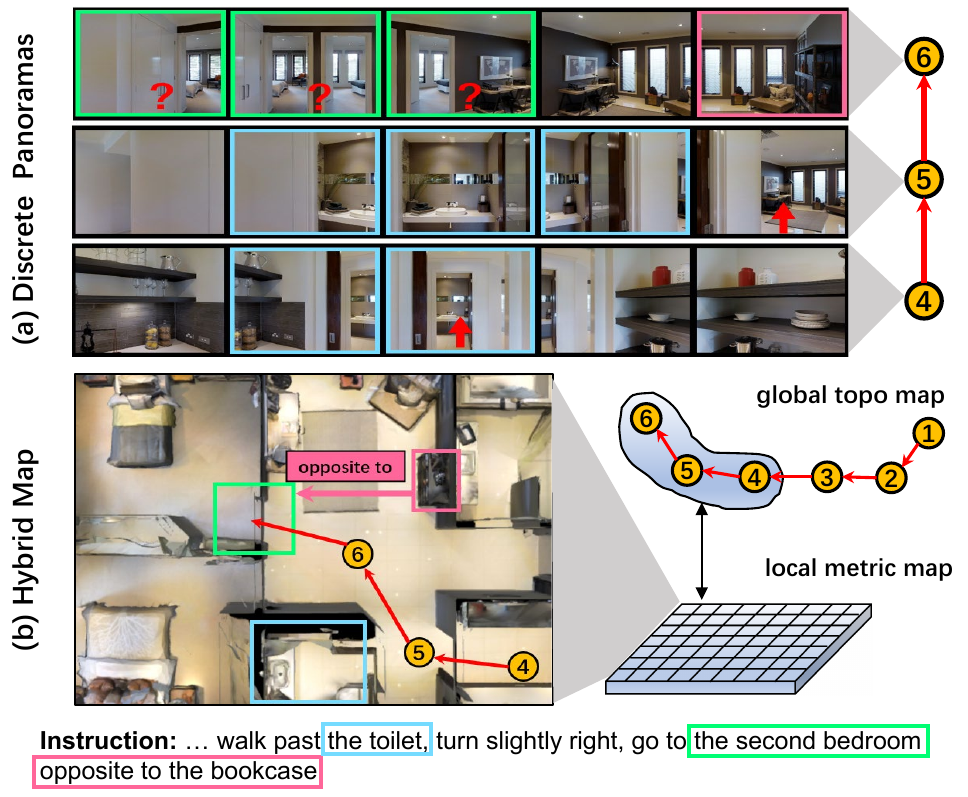}
\vspace{-3mm}
\caption{
(a) Incomplete observations within a single view and duplicates across views may confuse the agent.
(b) Projecting discrete panoramas into a unified map can solve the problem, thus facilitating spatial reasoning.
}\label{fig:intro1}
\vspace{-5mm}
\end{figure}
\label{sec:intro}
Interaction with an assistant robot using natural language is a long-standing goal. Towards this goal, vision-and-language navigation (VLN) has been proposed and drawn increasing research interest~\cite{anderson2018vision,qi2020reverie,ku2020room}.
Given a natural language instruction, a VLN agent is required to interpret and follow the instruction to reach the desired location. 
Enhancing the learning of visual-textual association is essential for the agent to succeed. 
Inspired by the great success of vision-language pre-training~\cite{chen2020uniter,tan2019lxmert,chen2023vlp,ji2023masked,gu2023eva2,wang2023large}, a variety of VLN pre-training methods have been studied and achieved promising results~\cite{hao2020towards,majumdar2020improving,guhur2021airbert,qi2021road,qiao2022hop}.

However, most existing VLN pre-training models resort to discrete panoramas (Fig.~\ref{fig:intro1} (a)) as visual inputs, which require the model to implicitly correlate incomplete, duplicate observations across views of the panoramas. This may hamper the agent's cross-modal spatial reasoning ability. 
As shown in Fig.~\ref{fig:intro1} (a), it is difficult to infer ``the second bedroom opposite to the bookcase'' because there are duplicate images of ``bedroom'' and ``bookcase'' across different views, and therefore it is hard to tell they are images for the same object or multiple instances.
A potential solution is to project these observations into a unified map, which explicitly aggregates incomplete observations and remove duplicate. 
Though this scheme has been successful in many navigation scenarios~\cite{chenweakly,chaplot2020object,chaplot2020neural}, 
its combination with pre-training remains unstudied, and this paper makes the first exploration.

In embodied navigation, maps generally fall into metric~\cite{henriques2018mapnet,chaplot2020object} or topological~\cite{chaplot2020neural,wang2021structured}. 
The metric map uses dense grid features to precisely describe the environment but has inefficiencies of scale~\cite{konolige2011navigation}. As a result, using a large map to capture the long-horizon navigation dependency can cause prohibitive computation~\cite{georgakis2022cross}, especially for the computation-intensive pre-training. Yet, such dependency has been shown crucial for VLN~\cite{chen2021history,qiao2022hop}.
On the other hand, the topo map can efficiently capture dependency by keeping track of visited locations as a graph structure~\cite{chaplot2020neural}. It also allows the agent to make efficient long-term goal plans, such as backtracking to a previous location~\cite{deng2020evolving,chen2022think}. However, each node in the graph is typically represented by condensed feature vectors, which lack fine-grained information for local spatial reasoning.

In this paper, instead of using a large global metric map, we propose a hybrid approach to balance the above two maps (shown in Fig.~\ref{fig:intro1} (b)). 
It contains a local metric map for short-term spatial reasoning while conducting overall long-term action plans on a global topo map.
This scheme shares similar spirits to classical topo-metric SLAM in robotics~\cite{blanco2008toward,konolige2011navigation}, but it differs in a learnable multimodal representation. 
To learn such representation, we propose BEVBert, a novel map-based pre-training paradigm that learns better visual-textual associations in bird's-eye view to aid complex spatial reasoning of VLN agents.
BEVBert first constructs offline hybrid maps based on large-scale VLN visual paths. 
Then, we employ a cross-modal transformer to conduct map-instruction interaction to obtain the multimodal map representation. 
To learn such representation, in addition to language modeling~\cite{kenton2019bert} and action prediction~\cite{hao2020towards}, we design a map prediction proxy task.
This task learns to encode linguistic and spatial priors to predict the information of unobserved regions, thereby reducing the uncertainty for decision-making. 
Finally, we fine-tune the model with sequential action prediction and online constructed hybrid maps. 
Thanks to the learned map representations, our agent learns a more robust navigation policy and achieves state-of-the-art on four VLN benchmarks (R2R, R2R-CE, RxR, REVERIE).

In summary, the contributions of this work are three-fold:

$\bullet$ We explore topo-metric maps in VLN for the first time. The proposed hybrid approach presents an elegant balance between short-term reasoning and long-term planning.

$\bullet$ We propose a novel map-based pre-training paradigm,  and empirically demonstrate that the learned map representation can enhance spatial-aware cross-modal reasoning.

$\bullet$ BEVBert achieves state-of-the-art on four VLN benchmarks (\eg, in test-unseen splits, 73 \texttt{SR} on R2R dataset, 59 \texttt{SR} on R2R-CE dataset, and 54.2 \texttt{SDTW} on RxR dataset).

%% file: 02_related.tex
\section{Related Work}
\label{sec:related}

\nbf{Vision-and-Language Navigation}
VLN has drawn increasing attention in recent years~\cite{anderson2018vision,qi2020reverie,krantz2020beyond,ku2020room,he2021landmark,gu2022vision,wang2022towards,Zhu2022diagnosing}.
Early VLN methods use sequence-to-sequence LSTMs to predict low-level actions~\cite{anderson2018vision} or high-level actions from discrete panoramas~\cite{fried2018speaker}. Different attention mechanisms~\cite{ma2019self,qi2020object,hong2020language,an2021neighbor} are proposed to improve cross-modal alignment. Reinforcement learning is also explored to enhance policy learning ~\cite{wang2018look,wang2019reinforced,tan2019learning}. 
To improve an agent's generalization ability to unseen environments, data augmentation strategies have been studied to mimic new environments~\cite{tan2019learning,liu2021vision,koh2021pathdreamer,li2022envedit,li2023improving,wang2023scale,wang2023lana,wang2022counterfactual,chen2022learning}. 
Recently, transformer-based models achieve good performance thanks to their powerful ability to learn generic multi-modal representations~\cite{hao2020towards,majumdar2020improving,guhur2021airbert}. This scheme is further extended by recurrent agent state~\cite{hong2021vln,qi2021road}, episodic memory~\cite{chen2021history,qiao2022hop}, or topology memory~\cite{zhao2022target,chen2022think,an2023etpnav} that significantly improves sequential action predictions. 
However, the widely used discrete panoramas~\cite{fried2018speaker} require implicit spatial modeling and may hamper the learning of generic language-environment correspondence. 
To address the limitation, we not only propose a multimodal topo-metric map but also devise a map-based pre-training framework.

\vspace{1mm}
\nbf{Visual Representation in Vision-Language Pre-training}
Existing approaches for VLP fall into image-based, object-based, and grid-based.
Image-based methods~\cite{radford2021learning} extract an overall feature for an image, yet neglect details, thus drawback on fine-grained language grounding.
Object-based methods~\cite{lu2019vilbert,tan2019lxmert} represent an image with dozens of objects identified by external detectors~\cite{anderson2018bottom,ren2015faster}. 
The challenge is that objects can be redundant and limited in predefined categories.
Grid-based methods~\cite{jiang2020defense,huang2021seeing} directly use image grid features for pre-training, thus enabling multi-grained vision-language alignments.
Most VLN pre-training are
image-based~\cite{majumdar2020improving,guhur2021airbert,qiao2022hop}, which rely on discrete panoramas.
We introduce grid-based methods into VLN through metric maps, where the model can learn via multi-grained room layouts.

\vspace{1mm}
\nbf{Maps for Navigation} 
Works on navigation have a long tradition of using SLAM~\cite{fuentes2015visual} to construct metric maps~\cite{chaplot2020object,narasimhan2020seeing}.
A metric map uses grid-based visual features to represent the scene layouts precisely but has inefficiencies of scale~\cite{konolige2011navigation}. To avoid heavy computation, standard practices restrict the map size\cite{georgakis2022cross,irshad2021sasra}, which can be inadequate for long-term modeling or planning.
Therefore, graph-based topo maps are proposed to address the limitation~\cite{chen2021topological,kwon2021visual,chaplot2020neural,wang2021structured}. But the drawback is short-term reasoning within the condensed nodes~\cite{chen2022think}.
In robotics, topo-metric maps are proposed to trade off their strengths~\cite{blanco2008toward,konolige2011navigation}. However, most of them are based on non-learning representations and focus on classical robotic tasks. 
We propose learnable topo-metric maps and explore the application to high-level VLN tasks.

%% file: 03_method.tex
\section{Method}
\label{sec:method}

The proposed method focuses on improving VLN agents' planning capability with map-based pre-training. 
For conciseness, we put our technical description in the context of VLN in discrete environments~\cite{anderson2018vision}, where maps can be derived from a predefined navigation graph. 
However, this method can also generalize to the task of VLN in continuous environments~\cite{krantz2020beyond} and more details are presented in \S~\ref{sec:sota}.

\vspace{2mm}
\nbf{Problem Definition}
An agent is required to follow an instruction $\mathbf{W}$ to traverse on a predefined graph $\mathbf{G}^{*}$ to reach the target location.
At each step $t$, the agent perceives a discrete panorama comprised of RGB images $\mathbf{V}_t$ and depth images $\mathbf{D}_t$.
Following~\cite{anderson2019chasing,deng2020evolving,wang2021structured,chen2022think}, we provide the agent with pose information $\mathbf{P}_t$ to simplify the mapping process. 
With observations $\mathbf{O}_t=\{\mathbf{V}_t,\mathbf{D}_t,\mathbf{P}_t\}$, VLN aims to learn a policy $\mathbf{\pi}(\mathbf{a}_t|\mathbf{W},\mathbf{O}_t)$ to predict action $\mathbf{a}_t$. The action is predicted by selecting a navigable node from a candidate set provided by the simulator. 
VLN datasets provide annotated instruction-path pairs to learn the policy, \ie, a pair contains an instruction $\mathbf{W}$ with $L$ words, and an expert path $\mathbf{\Gamma}=\langle \mathbf{O}_1,...,\mathbf{O}_T \rangle$ of length $T$.

\vspace{3mm}
\nbf{Method Overview}
As depicted in Fig.~\ref{fig:pipeline}, our map-based pre-training framework consists of two modules, namely topo-metric mapping and multimodal map learning.
The mapping module constructs an offline hybrid map via a sampled expert path (\S~\ref{sec:map}). 
The learning module conducts map-instruction interaction (\S~\ref{sec:arch}), and then learns multimodal map representations with three pre-training tasks (\S~\ref{sec:pretrain}).
After pre-training, the same model is fine-tuned on a sequential action prediction task with online constructed maps (\S~\ref{sec:train}).

\subsection{Topo-Metric Mapping} \label{sec:map}
\begin{figure*}[!tbp]
\centering
\includegraphics[width=\textwidth]{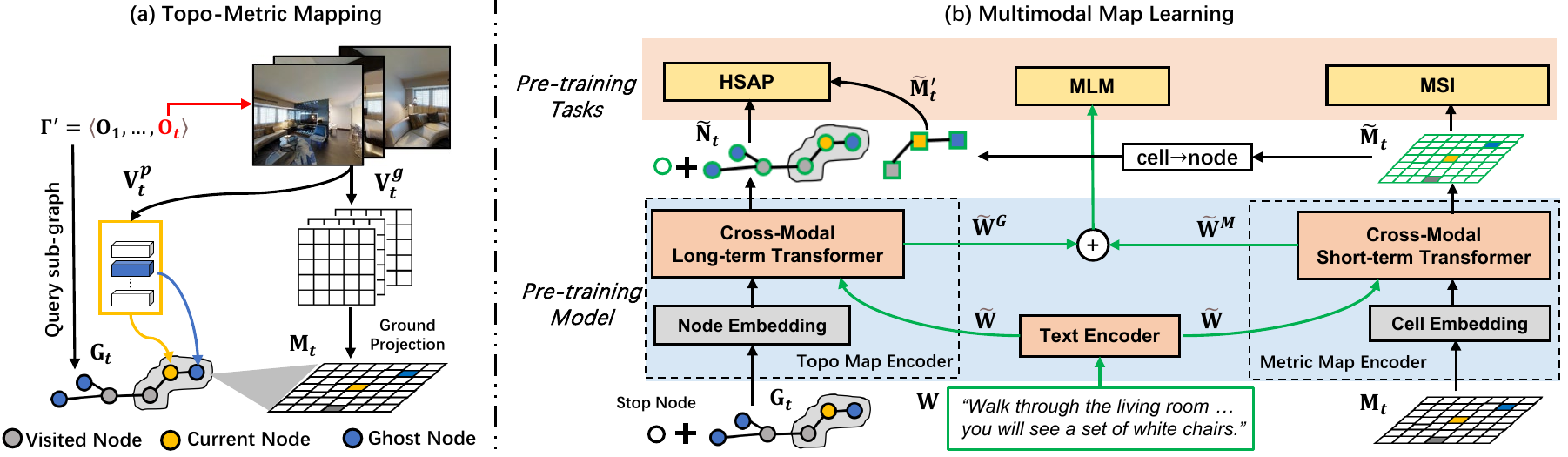}
\caption{The main architecture of the proposed hybrid-map-based pre-training framework.
}\label{fig:pipeline}
\end{figure*}

To balance the demand of VLN for long-term planning and short-term reasoning, we propose to construct a hybrid map.
As shown in Fig.~\ref{fig:pipeline} (a), assuming the agent currently is at step $t$ and the walked path is $\mathbf{\Gamma'}$, we construct a global topo map $\mathbf{G}_t$ and a local metric map $\mathbf{M}_t$. We next introduce how to construct these two maps. 

\vspace{1mm}
\nbf{Image Processing}
For panoramic RGB images $\mathbf{V}_t$ of each step $t$, we use a pre-trained vision transformer (ViT)~\cite{dosovitskiy2020image} to extract feature vectors $\mathbf{V}_{t}^p$ and downsized grid features $\mathbf{V}_{t}^g$. The associated depth images $\mathbf{D}_{t}$ are downsized to the same scale as $\mathbf{D'}_{t}$.

\vspace{1mm}
\nbf{Topo Mapping}
The graph-based topo map $\mathbf{G}_t=\{\mathbf{N}_t,\mathbf{E}_t\}$ keeps track of all observed nodes along the path $\mathbf{\Gamma'}$. 
Given $\mathbf{\Gamma'}$, we initialize $\mathbf{G}_t$ by deriving its corresponding sub-graph from the predefined graph $\mathbf{G}^*$. 
The nodes $\mathbf{N}_t$ are divided into three categories: visited node~\inlinegraphics{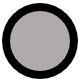}, current node~\inlinegraphics{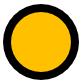} and ghost node~\inlinegraphics{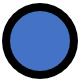}, where `ghost' denotes navigable nodes observed along the path $\mathbf{\Gamma'}$ but have not been explored. 
The edges $\mathbf{E}_t$ record the Euclidean distances among all adjacent nodes. 
We map feature vectors $\mathbf{V}_{*}^p$ onto the nodes as their visual representations. 
Taking time step $t$ as an example, $\mathbf{V}_{t}^p$ are first fed into a pano encoder~\cite{chen2021history} (a two-layer transformer) to obtain contextual view embeddings $\widehat{\mathbf{V}}_{t}^p$. 
Since \inlinegraphics{fig/current_node.png} and \inlinegraphics{fig/visited_node.png} have been visited and can access panoramas, they are represented by an average of panoramic view embeddings, \eg, $\textrm{Average}(\widehat{\mathbf{V}}_{t}^p) \in \mathcal{R}^D$ for \inlinegraphics{fig/current_node.png} ($D$ is the embedding dimension).
\inlinegraphics{fig/unexplored_node.png} is partially observed and therefore is represented by accumulated embeddings of views from which \inlinegraphics{fig/unexplored_node.png} can be observed. 
We equip $\mathbf{G}_t$ with a global action space $\mathcal{A}^G$ for long-term planning, which consists of all observed nodes. 

\vspace{1mm}
\nbf{Metric Mapping}
The grid-based metric map $\mathbf{M}_t \in \mathcal{R}^{U \times V \times D}$ is constructed locally centered on the current node~\inlinegraphics{fig/current_node.png}.
We define $\mathbf{M}_t$ as an egocentric map in which each cell contains a $D$-sized latent feature representing a small region of the surrounding layouts.
Similar to MapNet~\cite{henriques2018mapnet}, we ground project grid visual features $\mathbf{V}_{*}^g$ onto the cells to represent the map. 
Since $\mathbf{M}_t$ is a local representation and can be observed from nearby visited nodes of the current node, we integrate grid features from nearby visited nodes to construct the map. 
Concretely, assuming the current node is $\mathbf{n}_i$, we first query the topo map $\mathbf{G}_t$ to get its nearby visited nodes within $\kappa$ order: $\mathcal{N}_{\kappa}=\{\mathbf{n}_j | \mathrm{order}(\mathbf{n}_i,\mathbf{n}_j) \leq \kappa\}$. 
Then, we combine the grid features $\mathbf{V}_{*}^g$ of nodes in $\mathcal{N}_{\kappa}$, and project them onto the ground plane (centered on the current node), using the corresponding depths $\mathbf{D'}_{*}$ and poses $\mathbf{P}_{*}$. 
The projected features are discretized into the 2D spatial grid $\mathbf{M}_t$, using elementwise average pooling to handle feature collisions in a cell.
We equip $\mathbf{M}_t$ with a local action space $\mathcal{A}^M$ for short-term reasoning, which consists of the current node and its adjacent nodes. We compute these nodes' coordinates on $\mathbf{M}_t$ by ground projecting their poses onto the map, namely `node$\rightarrow$cell'.

\subsection{Pre-training Model} \label{sec:arch}
As presented in Fig.~\ref{fig:pipeline} (b), we then fed the hybrid map $(\mathbf{G}_t, \mathbf{M}_t)$ obtained in \S~\ref{sec:map} into a pre-training model to obtain multimodal map representations. 
The pre-training model contains a topo map encoder and a metric map encoder, which take the instruction $\mathbf{W}$ to fuse with $\mathbf{G}_t$ and $\mathbf{M}_t$ respectively. 
The outputs are later fed into three pre-training tasks to learn navigation-oriented multimodal map representations (\S~\ref{sec:pretrain}).

\vspace{-4mm}
\subsubsection{Text Encoder}
\vspace{-2mm}
Each word embedding in the instruction $\mathbf{W}$ is added with a position embedding~\cite{kenton2019bert} and a text type embedding~\cite{tan2019lxmert}.
Then, all embeddings are fed into a multi-layer transformer to obtain contextual word embeddings $\widetilde{\mathbf{W}}$.

\vspace{-4mm}
\subsubsection{Topo Map Encoder}~\label{sec:topo_encoder}
This module takes the topo map $\mathbf{G}_t$ and the encoded instruction $\widetilde{\mathbf{W}}$ to conduct node-level cross-modal fusion. 

\vspace{1mm}
\nbf{Node Embedding}
Each node feature $\mathbf{n}_i \in \mathbf{N}_t$ is added with a location embedding and a navigation step embedding.
The location embedding is calculated by the relative orientation and euclidean distance of each node to the current node, and the step embedding is the latest visited time step for visited nodes (\inlinegraphics{fig/visited_node.png}, \inlinegraphics{fig/current_node.png}) and 0 for ghost nodes~\inlinegraphics{fig/unexplored_node.png}. 
We add a zero-vector `stop' node $\mathbf{n}_{0}$ in the graph to denote a stop action and connect it with all other nodes. 

\nbf{Cross-modal Long-term Transformer}
The encoded node and word embeddings are fed into a multi-layer transformer to conduct node-level cross-modal fusion. 
The architecture of each layer is similar to LXMERT~\cite{tan2019lxmert}, which contains one bi-directional cross-attention sub-layer, two self-attention sub-layers, and two feed-forward sub-layers. 
Following \cite{chen2022think}, we replace the vision self-attention sub-layers with graph-aware self-attention (GASA), which introduces graph topology for node encoding.
The outputs are node-instruction-associated representations ($\widetilde{\mathbf{N}}_t$, $\widetilde{\mathbf{W}}^G$).

\vspace{-3mm}
\subsubsection{Metric Map Encoder}~\label{sec:metric_encoder}
This module takes the metric map $\mathbf{M}_t$ and the encoded instruction $\widetilde{\mathbf{W}}$ to conduct cell-level cross-modal fusion. 

\vspace{1mm}
\nbf{Cell Embedding}
Each cell feature $\mathbf{m}_{u,v} \in \mathbf{M}_t$ is added with a position embedding $\mathbf{p}_{u,v}$ 
and a navigability embedding $\mathbf{n}_{u,v}$.
To capture the relations between the agent and surrounding room layouts, we design an egocentric polar position embedding for each cell:
\begin{equation}
\vspace{-1mm}
\mathbf{p}_{u,v}=[\textrm{cos}(\mathbf{\theta}_{u,v}), \textrm{sin}(\mathbf{\theta}_{u,v}), \textrm{dis}_{u,v}]
\vspace{-1mm}
\end{equation}
where $\mathbf{\theta}_{u,v}$ and $\textrm{dis}_{u,v}$ denote the relative heading and normalized distance of a cell to the map center (agent position).
We empirically found it is better than a learnable~\cite{kenton2019bert} or 2D position embedding~\cite{dosovitskiy2020image}.
Navigability embeddings are set to 1 for cells that lie in the local action space $\mathcal{A}^M$, and 0 otherwise.
Both position and navigability embeddings are linearly transformed to $D$-dimension.

\nbf{Cross-modal Short-term Transformer}
The encoded cell and word embeddings are fed into a multi-layer transformer to conduct cross-modal fusion. Each layer architecture is similar to that in \S~\ref{sec:topo_encoder}, but uses self-attention for cell encoding rather than GASA. 
The short-term transformer conduct cross-modal reasoning on the fine-grained (cell-level) map representation, which can benefit reasoning about complicated spatial relations, such as \textit{``go into the hallway second to the right from the stairs"}.
The outputs are cell-instruction-associated representations ($\widetilde{\mathbf{M}}_t$, $\widetilde{\mathbf{W}}^M$).

\subsection{Pre-training Tasks}\label{sec:pretrain}
We devise three tasks to learn the multimodal map representations $(\widetilde{\mathbf{N}}_t, \widetilde{\mathbf{M}}_t)$ obtained in \S~\ref{sec:arch}.

\vspace{1mm}
\nbf{Masked Language Modeling (MLM)}
MLM is the most commonly used proxy task in BERT pre-training~\cite{kenton2019bert}. 
For VLN, MLM aims to recover masked words $\mathbf{W}_{m}$ via reasoning over the surrounding words $\mathbf{W}_{\setminus {m}}$ and the hybrid map.
Precisely, we first randomly mask out input tokens of the instruction with a 15\% probability and then conduct map-instruction interaction as explained in \S~\ref{sec:arch}.
To learn both long-term and short-term reasoning, we sum the obtained $\widetilde{\mathbf{W}}_{\setminus {m}}^G$ and $\widetilde{\mathbf{W}}_{\setminus {m}}^M$, then feed it into the MLM head. This task is optimized by minimizing the negative log-likelihood:
{\small
\begin{equation}
\vspace{-1mm}
\mathcal{L}_{\textrm{MLM}}
= -\mathbb{E}_{(\mathbf{W}, \mathbf{\Gamma'}) \sim \mathcal{D}} \log 
\mathcal{P}_{\mathbf{\theta}}(\mathbf{W}_{m} | \mathbf{W}_{\setminus m}, \mathbf{G}_t, \mathbf{M}_t)
\end{equation}
}%
where $\mathcal{D}$ denotes the training dataset and $\theta$ represents trainable parameters.

\vspace{1mm}
\nbf{Hybrid Single Action Prediction (HSAP)}
HSAP is designed to benefit the downstream goal: predicting navigation actions. 
Our model predicts an overall action in the global action space $\mathcal{A}^G$. 
For a more robust action plan, we integrate the short-term reasoning results from the metric map into the topo map.
In practice, we first convert cells lying in the local action space $\mathcal{A}^M$ into the global action space $\mathcal{A}^G$, using a `cell$\rightarrow$node' operation (the inverse of `node$\rightarrow$cell' in \S~\ref{sec:map}).
We denote the converted cells as $\widetilde{\mathbf{M'}}_{t} = \{ \tilde{\mathbf{m}}_i | i \in \mathcal{A}^{G'} \}$, where $\mathcal{A}^{G'}$ is a subset of the global action space $\mathcal{A}^{G}$.
Then, we use two feedforward networks (FFN) to predict navigation scores for nodes $\tilde{\mathbf{n}}_i \in \widetilde{\mathbf{N}}_t$ and cells $\tilde{\mathbf{m}}_i \in \widetilde{\mathbf{M'}}_t$, and dynamic fuse them conditioned on the agent state:
{\small
\begin{equation}\label{eq:action_score}
\vspace{-1mm}
\mathbf{s}_{i}^G = \textrm{FFN}(\tilde{\mathbf{n}}_i), \quad \mathbf{s}_{i}^M = \textrm{FFN}(\tilde{\mathbf{m}}_i)
\vspace{-1mm}
\end{equation}
}%
{\small
\begin{equation}\label{eq:action_fuse}
\mathbf{s}_i =\left\{
\begin{aligned}
& \mathbf{\delta}_t \mathbf{s}_{i}^G + (1-\mathbf{\delta}_t) \mathbf{s}_{i}^M,
~\textrm{if}~i \in \mathcal{A}^G \cap \mathcal{A}^{G'} \\
& \mathbf{s}_{i}^G,~\text{otherwise}
\end{aligned}
\right.
\end{equation}
}%
where the padded 'stop' node $\tilde{\mathbf{n}}_0$ and central cell $\tilde{\mathbf{m}}_{c,c}$ denote the agent state, therefore $\delta_t = \textrm{Sigmoid}(\textrm{FFN}([\tilde{\mathbf{n}}_0;\tilde{\mathbf{m}}_{c,c}])$.
In most VLN tasks, it is not necessary for an agent to revisit a node, therefore we mask the scores of visited nodes.
The task is optimized via a cross-entropy loss over fused scores $\{ \mathbf{s}_i \}$ and teacher action $\mathbf{a}_{t}^*$:
{\small
\begin{equation}\label{eq:hsap}
\vspace{-1mm}
\mathcal{L}_{\textrm{HSAP}} = -\mathbb{E}_{(\mathbf{W}, \mathbf{\Gamma}, {\mathbf{a}_{t}^*})\sim \mathcal{D}}
\log \mathcal{P}_{\theta}(\mathbf{a}_{t}^* | \mathbf{W}, \mathbf{G}_t, \mathbf{M}_t)
\end{equation}
}%

\nbf{Masked Semantic Imagination (MSI)}
We note there are some unobserved areas on the metric map $\mathbf{M}_t$, which brings uncertainty for decision-making.  
To mitigate this issue, we propose MSI to enable the agent to imagine the information of unobserved areas, by reasoning over instructions and partially observed maps. 
Concretely, we first randomly mask out cells of the metric map $\mathbf{M}_t$ with an empirical 15\% probability to simulate unobserved areas. 
Then, the masked map $\mathbf{M}_{t,\setminus m}$ interacts with the instruction $\mathbf{W}$ as explained in \S~\ref{sec:arch}. 
Finally, the MSI head forces the model to predict semantics $\mathbf{S}$ of masked regions conditioned on the multimodal map representation $\widetilde{\mathbf{M}}_{t,\setminus m}$.
Each cell of the metric map may contain multiple semantics; therefore, the task is formulated as a multi-label classification problem and optimized via a binary cross-entropy loss:
{\small
\begin{equation}
\vspace{-2mm}
\begin{aligned}
\mathcal{L}_{\textrm{MSI}}=-\mathbb{E}_{(\mathbf{W}, \mathbf{\Gamma}) \sim \mathcal{D}} \sum_{i}^C 
[ \mathbf{S}_{i} \log \mathcal{P}_{\theta}(\mathbf{S}_{i} | \mathbf{W}, \mathbf{M}_{t,\setminus m}) \\
+ (1-\mathbf{S}_{i}) \log (1 - \mathcal{P}_{\theta}(\mathbf{S}_{i} | \mathbf{W}, \mathbf{M}_{t,\setminus m})) ]
\end{aligned}
\end{equation}
}%
where $\mathbf{S}_{i}$ corresponds to the $i$-th semantic class ($C=40$), and we obtain these labels from Matterport3D dataset~\cite{chang2017matterport3d}.

\subsection{Training and Inference} \label{sec:train}
\begin{figure}[!htbp]
\centering
\includegraphics[width=0.48\textwidth]{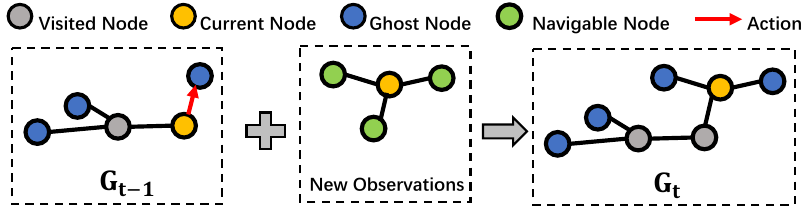}
\vspace{-6mm}
\caption{
Online topo map update at step $t$. The agent executes an action to reach a ghost node and receives new observations. It then adds newly observed nodes to $\mathbf{G}_{t-1}$, updating node representations and types. The simulator provides navigable nodes at each step. 
}\label{fig:online_map}
\vspace{-3mm}
\end{figure}

\nbf{Training}
As standard practices in transformer-based VLN methods~\cite{hao2020towards,majumdar2020improving,guhur2021airbert}, we first mix the three tasks in \S~\ref{sec:pretrain} to pre-train the model with offline expert data. 
To avoid overfitting to expert experience, we then fine-tune the model with sequential action prediction.
The topo map $\mathbf{G}_t$ in this stage is online updated.
As shown in Fig.~\ref{fig:online_map}, at step $t$, we obtain $\mathbf{G}_t$ by adding newly observed nodes to $\mathbf{G}_{t-1}$ and updating the node status. 
For trajectory rollout in fine-tuning, we alternately run `teacher-forcing' and `student-forcing'~\cite{anderson2018vision}.
The `teacher-forcing' is equivalent to Eq.~\ref{eq:hsap}, where the agent always executes the teacher action. 
In 'student-forcing', at each step, the next action is sampled from the predicted score distribution (Eq.~\ref{eq:action_fuse}) and supervised by pseudo labels~\cite{chen2022think}. More details are in the appendix.

\nbf{Inference}
At each step during testing, the agent online constructs a hybrid map similar to the fine-tuning stage, and then performs cross-modal reasoning over the map as explained in \S~\ref{sec:arch}. 
Following the single-run setting of VLN, the agent greedily selects the node (ghost node or `stop' node) with the maximum predicted score (Eq.~\ref{eq:action_fuse}) as the next action. 
If the selected node is a long-term action (not adjacent to the current node), the agent plans the shortest path to reach the selected node using Dijkstra's algorithm on the current topo map. 
The agent stops if it selects the `stop' node or reaches the maximum action steps.

%% file: 04_experiments.tex
\section{Experiments}
\label{sec:exprs}
We evaluate the proposed method on R2R~\cite{anderson2018vision}, R2R-CE~\cite{krantz2020beyond}, RxR~\cite{ku2020room} and REVERIE~\cite{qi2020reverie} datasets. 
R2R, R2R-CE, and RxR focus on fine-grained instruction following, whereas R2R-CE is a variant of R2R in continuous environments and RxR provides more detailed path descriptions (\eg, objects and their relations).
REVERIE is a goal-oriented task using coarse-grained instructions, such as ``Go to the entryway and clean the coffee table''.

\nbf{Navigation Metrics} 
As in~\cite{anderson2018vision,anderson2018evaluation,ilharco2019general}, we adopt the following navigation metrics. Trajectory Length (\texttt{TL}): average path length in meters; Navigation Error (\texttt{NE}): average distance in meters between the final and target location; Success Rate (\texttt{SR}): the ratio of paths with \texttt{NE} less than 3 meters; Oracle \texttt{SR} (\texttt{OSR}): \texttt{SR} given oracle stop policy; \texttt{SR} penalized by Path Length (\texttt{SPL}); Normalize Dynamic Time Wrapping (\texttt{NDTW}): the fidelity between the predicted and annotated paths and \texttt{NDTW} penalized by \texttt{SR} (\texttt{SDTW}).

\nbf{Object Grounding Metrics}
As in~\cite{qi2020reverie}, we use Remote Grounding Success (\texttt{RGS}) and \texttt{RGSPL} (\texttt{RGS} penalized by Path Length) to evaluate the capacity of object grounding.
All metrics are the higher the better, except for \texttt{TL} and \texttt{NE}.

\subsection{Implementation Details}
\nbf{Image Processing and Mapping} 
We resize and central crop RGB images to $224\times224$. 
Following~\cite{shen2021much,li2022envedit}, we use ViT-B/16-CLIP~\cite{radford2021learning} to extract visual features. The scale of grid visual features $\mathbf{V}_{t}^{g}$ is $14\times14$ (outputs before the MLP head of ViT).
We set the metric map scale as $21 \times 21$, and each cell represents a square region with a side length of 0.5m (the entire map is thus $10.5m \times 10.5m$).

\vspace{1mm}
\nbf{Model Configuration}
Following~\cite{hong2021vln,chen2022think}, we set the layers' number of the text encoder, and the two map encoders as 9, 4, 4.  
Other hyperparameters are the same as LXMERT~\cite{tan2019lxmert} (\eg the hidden layer size is 768). 
In the pre-training stage, we use pre-trained LXMERT for initialization on R2R, R2R-CE, and REVERIE datasets, and pre-trained RoBerta~\cite{liu2019roberta} is used for the multilingual RxR dataset.
REVERIE provides additional object annotations for the final object grounding task, and BEVBert adaptation to this dataset is presented in the appendix.

\vspace{1mm}
\nbf{Training Details} 
The trainable modules in our model include the pano encoder in \S~\ref{sec:map}, the text encoder, and the two map encoders. 
For all datasets, we first offline pre-train BEVBert with batch size 64 for 100k iterations using 4 NVIDIA Tesla A100 GPUs ($\sim$10 hours). 
We use the Prevalent~\cite{hao2020towards}, RxR-Markey~\cite{wang2022less} and REVERIE-Spk~\cite{chen2022think} synthetic instructions as data augmentation on R2R/R2R-CE, RxR and REVERIE respectively. 
We choose a pre-trained model with the best zero-shot performance (\eg, \texttt{SR} + \texttt{SPL} on R2R/R2R-CE, \texttt{SR} + \texttt{NDTW} on RxR, \texttt{SR} + \texttt{RGS} on REVERIE) as initialization for downstream fine-tuning.
Then, we use alternative teacher-forcing and student-forcing to online fine-tune the model in the simulator, with batch size 16 for 40k iterations on 4 NVIDIA Tesla A100 GPUs ($\sim$20 hours). 
The best iterations are selected by best performance on validation unseen splits.

\subsection{Comparison with State-of-the-Art}\label{sec:sota}

\nbf{R2R}
Tab.~\ref{tab:r2r} compares BEVBert against state-of-the-art (SoTA) methods on the R2R dataset.
BEVBert beats other methods on all evaluation metrics except for the ensemble-based EnvEdit~\cite{li2022envedit}.
On the test unseen split, for instance, BEVBert outperforms the previous best method DUET~\cite{chen2022think} by 4 \texttt{SR} and 3 \texttt{SPL}.
It is worth noticing that compared with Chasing~\cite{anderson2019chasing} which also uses metric maps, our improvement is substantial ($\uparrow$ 40 \texttt{SR} and $\uparrow$ 32 \texttt{SPL} on the test unseen split).
We attribute this to our hybrid map design, which balances short-term reasoning and long-term planning, whereas Chasing resorts to metric maps, leading to non-ideal long-term planning capacity. 
Moreover, Chasing is trained from scratch, while BEVBert gains superior generalization ability with the proposed pre-training framework. 

\vspace{1mm}
\nbf{R2R-CE}
Tab.~\ref{tab:r2r_ce} presents the results on the R2R-CE dataset. 
We adjust the topo mapping process in \S~\ref{sec:map} to adapt BEVBert to continuous environments. 
Specifically, at each step, the agent predicts a set of waypoints~\cite{hong2022bridging} and organizes them as a topo map similar to~\cite{an2023etpnav}. 
BEVBert sets new SoTA on the R2R-CE dataset, with 4 \texttt{SR} and 2 \texttt{SPL} improvement over the topo-map-only ETPNav~\cite{an2023etpnav}.
This further highlights the efficacy of the proposed hybrid map.

\vspace{1mm}
\nbf{RxR}
Tab.~\ref{tab:rxr} reports the results on the RxR dataset. 
RxR is more challenging than R2R because its paths are much longer and involve more detailed path descriptions. 
With the fine-grained metric map, BEVBert is skilled at these complex instructions and achieves considerable improvement. 
For instance, on the test unseen split, BEVBert surpasses the ensemble-based EnvEdit~\cite{li2022envedit} by 4 \texttt{SR}, 0.8 \texttt{NDTW} and 2.4 \texttt{SDTW}.
We also report BEVBert's performance without Marky synthetic instructions~\cite{wang2022less}. 
Compared to EnvEdit, BEVBert still leads on \texttt{SR}, and the improvements over the SoTA single-model HAMT~\cite{chen2021history} are notable (\eg $\uparrow$ 7.6 \texttt{SR}, $\uparrow$ 0.8 \texttt{NDTW} and $\uparrow$ 4.3 \texttt{SDTW} on the val unseen split).

\definecolor{Gray}{gray}{0.9}
\begin{table}[htbp]
\centering
\resizebox{0.48\textwidth}{!}{\begin{tabular}{l cccc | cccc}
\toprule 
\multicolumn{1}{c}{} & \multicolumn{4}{c}{Val Unseen} & \multicolumn{4}{c}{Test Unseen} \\
\cmidrule(r){2-5} \cmidrule(r){6-9}
\multicolumn{1}{c}{Methods} & 
\multicolumn{1}{c}{NE$\downarrow$} & \multicolumn{1}{c}{OSR$\uparrow$} & \multicolumn{1}{c}{\textbf{SR}$\uparrow$} & \multicolumn{1}{c}{\textbf{SPL}$\uparrow$} & 
\multicolumn{1}{c}{NE$\downarrow$} & \multicolumn{1}{c}{OSR$\uparrow$} & \multicolumn{1}{c}{\textbf{SR}$\uparrow$} & \multicolumn{1}{c}{\textbf{SPL}$\uparrow$} \\
\midrule
Seq2Seq~\cite{anderson2018vision}
& 7.81 & 28 & 21 & - 
& 7.85 & 27 & 20 & - \\
SF~\cite{fried2018speaker}
& 6.62 & 45 & 36 & - 
& 6.62 & - & 35 & 28  \\
Chasing~\cite{anderson2019chasing}
& 7.20 & 44 & 35 & 31
& 7.83 & 42 & 33 & 30 \\
RCM~\cite{wang2019reinforced}
& 6.09 & 50 & 43 & - 
& 6.12 & 50 & 43 & 38 \\
SM~\cite{ma2019self}
& 5.52 & 56 & 45 & 32 
& 5.67 & 59 & 48 & 35 \\
EnvDrop~\cite{tan2019learning}
& 5.22 & - & 52 & 48
& 5.23 & 59 & 51 & 47 \\
AuxRN~\cite{zhu2020vision}
& 5.28 & 62 & 55 & 50
& 5.15 & 62 & 55 & 51 \\
NvEM~\cite{an2021neighbor}
& 4.27 & - & 60 & 55
& 4.37 & 66 & 58 & 54 \\
SSM~\cite{wang2021structured}
& 4.32 & 73 & 62 & 45
& 4.57 & 70 & 61 & 46 \\
PREVAL~\cite{hao2020towards}\dag
& 4.71 & - & 58 & 53
& 5.30 & 61 & 54 & 51 \\
AirBert~\cite{guhur2021airbert}\dag
& 4.10 & - & 62 & 56 
& 4.13 & - & 62 & 57 \\
RecBert~\cite{hong2021vln}\dag
& 3.93 & - & 63 & 57
& 4.09 & 70 & 63 & 57 \\
REM~\cite{liu2021vision}
& 3.89 & - & 64 & 58
& 3.87 & 72 & 65 & 59 \\
HAMT~\cite{chen2021history}\dag
& 3.65 & - & 66 & 61
& 3.93 & 72 & 65 & 60 \\
HOP+~\cite{qiao2023hop+}\dag
& 3.49 & - & 67 & 61
& 3.71 & - & 66 & 60 \\
EnvEdit*~\cite{li2022envedit}\dag
& 3.24 & - & 69 & \color{blue}64 
& 3.59 & - & 68 & \color{blue}64 \\
TD-STP~\cite{zhao2022target}\dag
& 3.22 & 76 & 70 & 63 
& 3.73 & 72 & 67 & 61 \\
DUET~\cite{chen2022think}\dag
& 3.31 & 81 & 72 & 60
& 3.65 & 76 & 69 & 59 \\
BEVBert (Ours)\dag
& \color{blue}2.81 & \color{blue}84 & \color{blue}75 & \color{blue}64
& \color{blue}3.13 & \color{blue}81 & \color{blue}73 & 62 \\
\bottomrule
\end{tabular}}
\vspace{-3mm}
\caption{Comparison with SoTA methods on R2R dataset. * Ensemble of three agents. \dag Pre-training-based methods.}\label{tab:r2r}
\vspace{-2mm}
\end{table}

\begin{table}[htbp]
\centering
\resizebox{0.48\textwidth}{!}{\begin{tabular}{l cccc | cccc}
\toprule 
\multicolumn{1}{c}{} & \multicolumn{4}{c}{Val Unseen} & \multicolumn{4}{c}{Test Unseen} \\
\cmidrule(r){2-5} \cmidrule(r){6-9}
\multicolumn{1}{c}{Methods} & 
\multicolumn{1}{c}{NE$\downarrow$} & \multicolumn{1}{c}{OSR$\uparrow$} & \multicolumn{1}{c}{\textbf{SR}$\uparrow$} & \multicolumn{1}{c}{\textbf{SPL}$\uparrow$} & 
\multicolumn{1}{c}{NE$\downarrow$} & \multicolumn{1}{c}{OSR$\uparrow$} & \multicolumn{1}{c}{\textbf{SR}$\uparrow$} & \multicolumn{1}{c}{\textbf{SPL}$\uparrow$} \\
\midrule
Seq2Seq~\cite{krantz2020beyond}
& 7.37 & 40 & 32 & 30
& 7.91 & 36 & 28 & 25 \\
CM2~\cite{georgakis2022cross}
& 7.02 & 42 & 34 & 28
& 7.70 & 39 & 31 & 24 \\
HPN~\cite{krantz2021waypoint}
& 6.31 & 40 & 36 & 34
& 6.65 & 37 & 32 & 30 \\
MGMAP~\cite{chenweakly}
& 6.28 & 48 & 39 & 34
& 7.11 & 45 & 35 & 28 \\
CWP~\cite{hong2022bridging}
& 5.74 & 53 & 44 & 39
& 5.89 & 51 & 42 & 36 \\
Sim2Sim~\cite{krantz2022sim}
& 6.07 & 52 & 43 & 36
& 6.17 & 52 & 44 & 37 \\
Reborn~\cite{an20221st}
& 5.40 & 57 & 50 & 46
& 5.55 & 57 & 49 & 45 \\
ETPNav~\cite{an2023etpnav}
& 4.71 & 65 & 57 & 49
& 5.12 & 63 & 55 & 48 \\
BEVBert (Ours)
& \color{blue}4.57 & \color{blue}67 & \color{blue}59 & \color{blue}50
& \color{blue}4.70 & \color{blue}67 & \color{blue}59 & \color{blue}50 \\
\bottomrule
\end{tabular}}
\vspace{-3mm}
\caption{Comparison with SoTA methods on R2R-CE dataset.}\label{tab:r2r_ce}
\vspace{-2mm}
\end{table}

\definecolor{Gray}{gray}{0.9}
\begin{table}[htbp]
\centering
\resizebox{0.48\textwidth}{!}{\begin{tabular}{l cccc | cccc}
\toprule 
\multicolumn{1}{c}{} & \multicolumn{4}{c}{Val Unseen} & \multicolumn{4}{c}{Test Unseen} \\
\cmidrule(r){2-5} \cmidrule(r){6-9}
\multicolumn{1}{c}{Methods} & 
\multicolumn{1}{c}{NE$\downarrow$} & \multicolumn{1}{c}{\textbf{SR}$\uparrow$} & \multicolumn{1}{c}{\textbf{NDTW}$\uparrow$} & \multicolumn{1}{c}{\textbf{SDTW}$\uparrow$} &
\multicolumn{1}{c}{NE$\downarrow$} & \multicolumn{1}{c}{\textbf{SR}$\uparrow$} & \multicolumn{1}{c}{\textbf{NDTW}$\uparrow$} & \multicolumn{1}{c}{\textbf{SDTW}$\uparrow$} \\
\midrule
LSTM~\cite{ku2020room}
& 10.9 & 22.8 & 38.9 & 18.2
& 12.0 & 21.0 & 36.8 & 16.9 \\
EnvDrop+~\cite{shen2021much}
& - & 42.6 & 55.7 & - 
& - & 38.3 & 51.1 & 32.4 \\
CLEAR-C~\cite{li2022clear}
& - & - & - & -
& - & 40.3 & 53.7 & 34.9 \\
HAMT~\cite{chen2021history}
& - & 56.5 & 63.1 & 48.3
& 6.2 & 53.1 & 59.9 & 45.2 \\
EnvEdit*~\cite{li2022envedit}
& - & 62.8 & 68.5 & 54.6 
& 5.1 & 60.4 & 64.6 & 51.8 \\
BEVBert\dag (ours)
& 4.6 & 64.1 & 63.9 & 52.6
& - & - & - & - \\
BEVBert (ours)
& \color{blue}4.0 & \color{blue}68.5 & \color{blue}69.6 & \color{blue}58.6
& \color{blue}4.8 & \color{blue}64.4 & \color{blue}65.4 & \color{blue}54.2 \\
\bottomrule
\end{tabular}}
\vspace{-3mm}
\caption{Comparison with SoTA methods on RxR dataset. * Ensemble of three agents. 
\dag Without Markey-T5 instructions~\cite{wang2022less}. }\label{tab:rxr}
\vspace{-2mm}
\end{table}

\begin{table}[htbp]
\centering
\resizebox{0.45\textwidth}{!}{\begin{tabular}{l ccc | ccc}
\toprule
\multicolumn{1}{c}{} & \multicolumn{3}{c}{Val Unseen} & \multicolumn{3}{c}{Test Unseen} \\
\cmidrule(r){2-4} \cmidrule(r){5-7}
\multicolumn{1}{c}{Methods}
& \multicolumn{1}{c}{\textbf{SR}$\uparrow$} & \multicolumn{1}{c}{\textbf{RGS}$\uparrow$} & \multicolumn{1}{c}{\textbf{RGSPL}$\uparrow$} 
& \multicolumn{1}{c}{\textbf{SR}$\uparrow$} & \multicolumn{1}{c}{\textbf{RGS}$\uparrow$} & \multicolumn{1}{c}{\textbf{RGSPL}$\uparrow$} \\ 
\midrule
AutoVLN*~\cite{chen2022learning}
& 55.89 & 36.58 & 26.76 
& 55.17 & 32.23 & 22.68 \\
\midrule
FAST~\cite{qi2020reverie} 
& 14.40 & 7.84 & 4.67 
& 19.88 & 11.28 & 6.08 \\
SIA~\cite{lin2021scene }
& 31.53 & 22.41 & 11.56 
& 30.80 & 19.02 & 9.20 \\
RecBert~\cite{hong2021vln} 
& 30.67 & 18.77 & 15.27 
& 29.61 & 16.50 & 13.51 \\
AirBert~\cite{guhur2021airbert}
& 27.89 & 18.23 & 14.18 
& 30.26 & 16.83 & 13.28 \\
HAMT~\cite{chen2021history} 
& 32.95 & 18.92 & 17.28 
& 30.40 & 14.88 & 13.08 \\ 
TD-STP~\cite{zhao2022target} 
& 34.88 & 21.16 & 16.56 
& 35.89 & 19.88 & 15.40 \\ 
DUET~\cite{chen2022think}
& 46.98 & 32.15 & 23.03 
& 52.51 & 31.88 & 22.06 \\
BEVBert (Ours)
& \color{blue}51.78 & \color{blue}34.71 & \color{blue}24.44 
& \color{blue}52.81 & \color{blue}32.06 & \color{blue}22.09 \\
\bottomrule
\end{tabular}}
\vspace{-3mm}
\caption{Comparison with SoTA methods on REVERIE dataset. * 900 extra scenes for training.}\label{tab:reverie}
\vspace{-4mm}
\end{table}

\vspace{1mm}
\nbf{REVERIE}
BEVBert also generalizes well on the goal-oriented REVERIE dataset as shown in Tab.~\ref{tab:reverie}.
On the val unseen split, BEVBert surpasses the previous best model DUET~\cite{chen2022think} by 4.80 \texttt{SR}, 2.56 \texttt{RGS}, and 1.41 \texttt{RGSPL}. 
We also note improvements on the test unseen split are less prominent compared to DUET.
We attribute it to the distribution shift between the val unseen and test unseen splits (\eg, comparing the performance difference between val unseen and test unseen, HAMT $\downarrow$ 2.55 \texttt{SR} \textit{v.s.} DUET $\uparrow$ 5.33 \texttt{SR}).

\subsection{Quantitative and Qualitative Analysis}
\label{sec:analysis}
We present quantitative and qualitative analyses to illustrate BEVBert's efficacy for complex spatial reasoning. 

\begin{figure}[!htbp]
\centering
\includegraphics[width=0.49\textwidth]{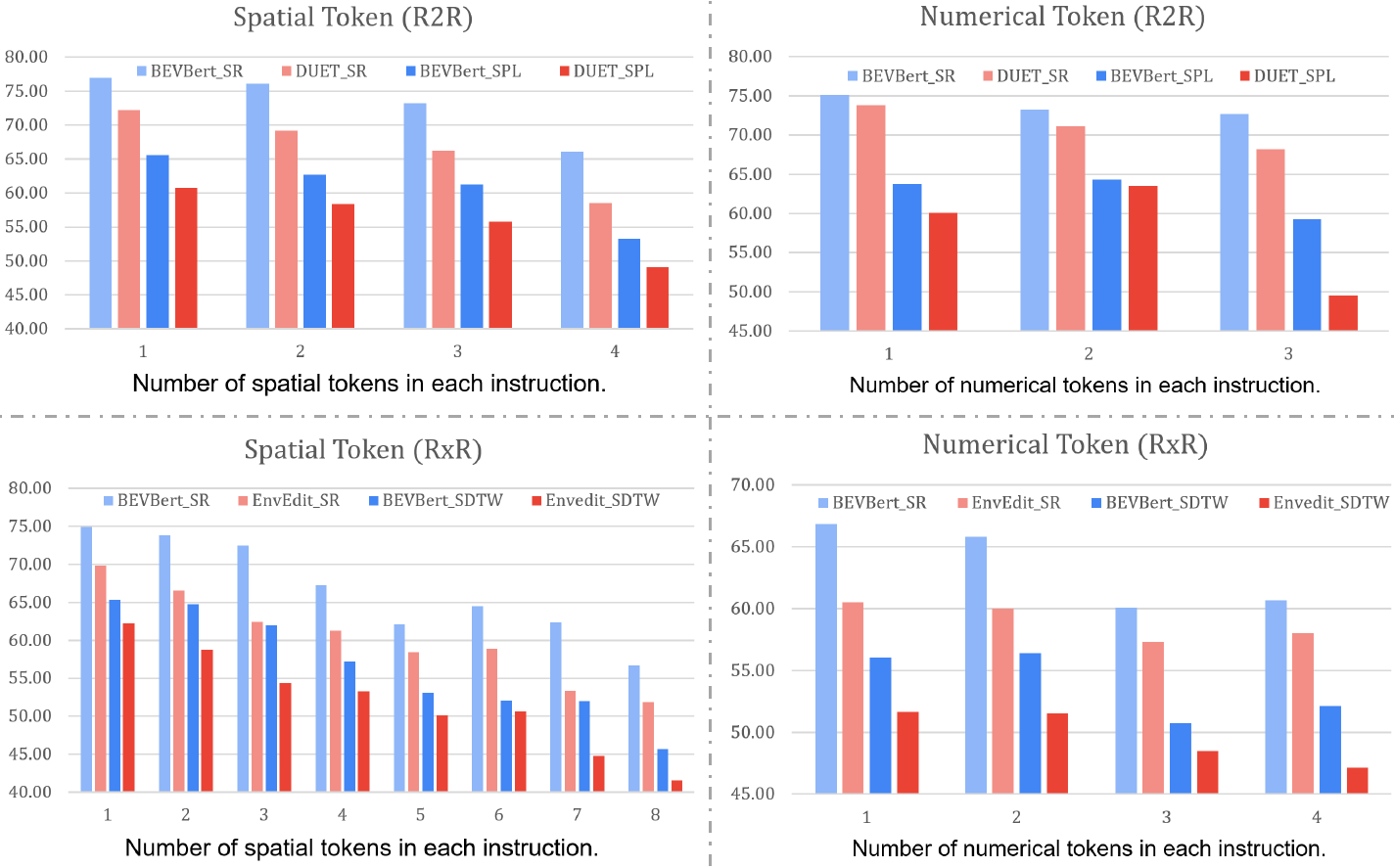}
\vspace{-6mm}
\caption{
Comparison of navigation performance on spatial and numerical related instructions 
({\color{blue}BEVBert} vs. {\color{red}DUET}~\cite{chen2022think} \texttt{SR} (light color) and \texttt{SPL} (dark color) on R2R val unseen split, 
{\color{blue}BEVBert} vs. {\color{red}EnvEdit}~\cite{li2022envedit} \texttt{SR} and \texttt{SDTW} on RxR val unseen split). 
}\label{fig:plot}
\vspace{-3mm}
\end{figure}

\nbf{Quantitative Analysis}
We aim to evaluate BEVBert's performance on instructions that involve spatial reasoning, such as ``go into the hallway second to the right from the stairs''.
Thus, from R2R and RxR val unseen splits, we first extract the relevant instructions which contain either spatial tokens (\eg ``left of'', ``rightmost'') or numerical tokens (\eg ``second'', ``fourth''). 
An agent's reasoning capability can be inferred from how well it follows these instructions. 
We compare the performance of BEVBert and SoTA methods on these instructions in Fig.~\ref{fig:plot}.
As the number of special tokens in each instruction increases, the performance of all models shows downward trends.
This indicates spatial reasoning is a bottleneck of existing methods.
However, BEVBert consistently outperforms these counterparts, especially on the RxR dataset which contains more spatial descriptions. 
This highlights BEVBert's superiority in spatial reasoning. 

\begin{figure}[!htbp]
\centering
\includegraphics[width=0.48\textwidth]{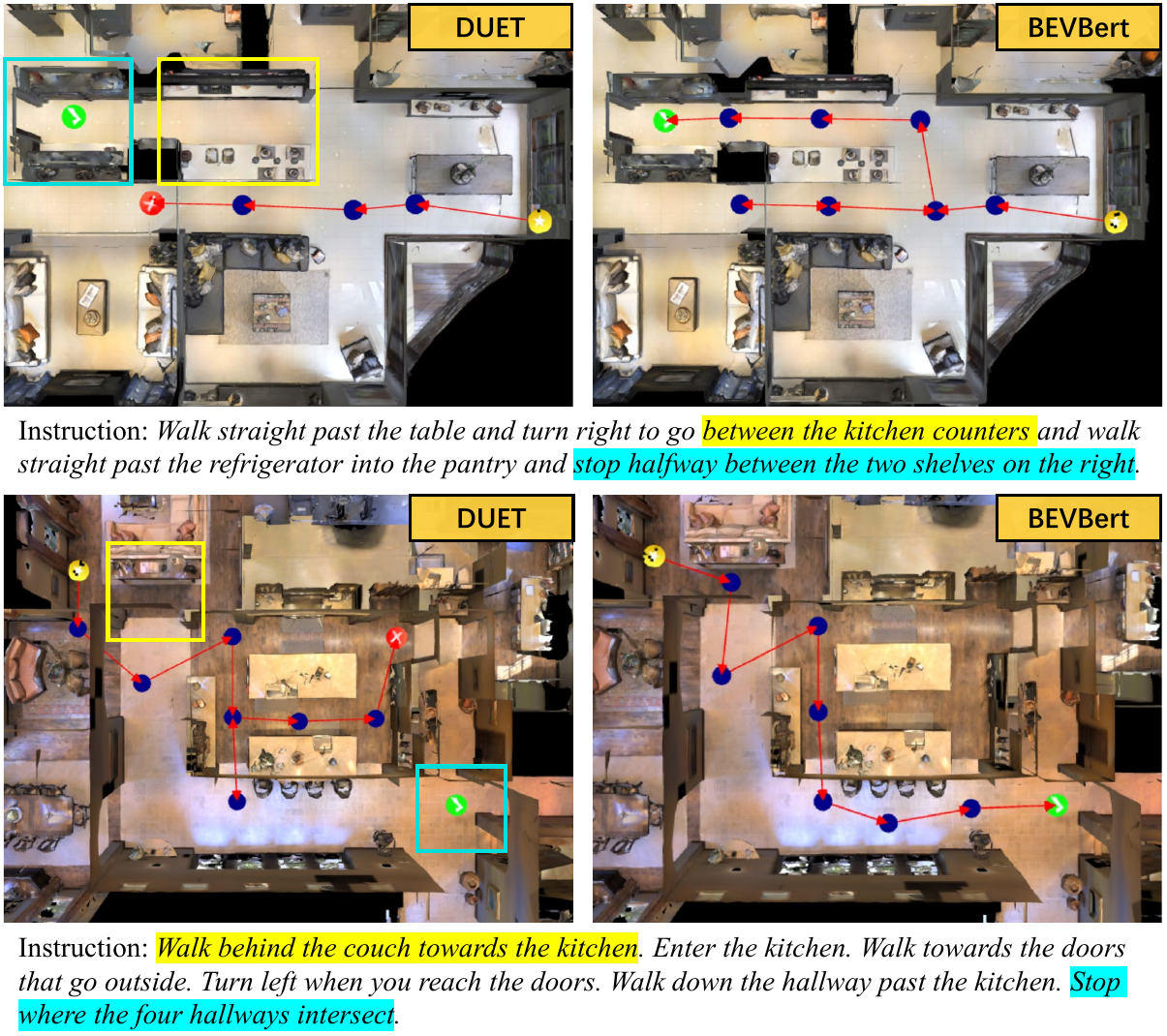}
\vspace{-6mm}
\caption{
Predicted paths of DUET~\cite{chen2022think} and BEVBert on R2R-unseen.
Yellow and green circles denote the start and target locations, respectively, and the red circles represent incorrect endpoints. 
}\label{fig:viz}
\vspace{-2mm}
\end{figure}

\nbf{Qualitative Analysis}
We visualize the predicted paths of BEVBert and DUET~\cite{chen2022think} in Fig.~\ref{fig:viz}. DUET uses discrete panoramas for local reasoning, leading to non-ideal spatial reasoning capacity. 
For example, it does not follow the instruction strictly (\eg ``go between the kitchen counters'', ``walk behind the couch'') and leads to incorrect endpoints. 
By contrast, thanks to the explicit spatial representation, BEVBert could interpret these complicated descriptions and make correct decisions.

\subsection{Ablation Study}
\label{sec:ablation}
We conduct extensive experiments to evaluate key design choices of BEVBert. 
Results are reported on the R2R val unseen split and the main metrics are highlighted.

\vspace{1mm}
\nbf{1) Comparison of map variants}
Tab.~\ref{tab:hybrid} presents the results of our model trained with different map variants. 
Row 1 only uses topo maps for action prediction. It achieves a decent 70.25 \texttt{SR}, but there is a clear gap ($\sim$ 4.5 \texttt{SR}) with hybrid maps (Row 5, Row 6), due to the lack of metric information for local spatial reasoning. 
Row 2 further fuses depth features~\cite{wijmansdd} into topo maps' node representations, but with no gain. This suggests that simple depth fusion cannot improve spatial reasoning ability. 
Row 3 and Row 4 only use metric maps, leading to higher \texttt{TL} but poorer navigation performance (\texttt{OSR} and \texttt{SR}), because the agent lacks long-term planning ability and makes some ineffective exploration. 
In Row 5 and Row 6, the navigation performance increases substantially when applying the proposed topo-metric maps.
It indicates that the proposed hybrid map is a good trade-off between the above two maps, which enables long-term and short-term balanced decision-making.

\definecolor{Gray}{gray}{0.9}
\begin{table}[!tbp]
\centering
\resizebox{0.45\textwidth}{!}{
\begin{tabular}{clc | ccccc}
\toprule
\multicolumn{1}{c}{\#} &
\multicolumn{1}{c}{Map} &
\multicolumn{1}{c}{Depth} & 
\multicolumn{1}{c}{TL} & \multicolumn{1}{c}{NE$\downarrow$} & \multicolumn{1}{c}{OSR$\uparrow$} & \multicolumn{1}{c}{\textbf{SR}$\uparrow$} & \multicolumn{1}{c}{\textbf{SPL}$\uparrow$} \\
\midrule
1 & \multirow{2}{*}{Topo} & - & 12.59 & 3.39 & 78.01 & 70.25 & 61.29 \\
2 & & sensing\dag  & 11.76 & 3.38 & 77.95 & 70.03 & 61.45 \\
\midrule
3 & \multirow{2}{*}{Metric} & estimated & 14.41 & 3.91 & 70.35 & 60.64 & 52.17 \\
4 &  & sensing   & 14.15 & 3.93 & 70.95 & 60.90 & 52.80 \\
\midrule
5 & \multirow{2}{*}{Hybrid} & estimated & 13.61 & 2.88 & 82.63 & 74.67 & \color{blue}63.63 \\
6 &  & sensing   & 14.55 & 2.81 & 83.65 & \color{blue}74.88 & 63.60 \\
\bottomrule
\end{tabular}}
\vspace{-2.5mm}
\caption{Comparison of map variants. We denote ground-truth and estimated depths as `sensing' and `estimated' respectively.
\dag~represents fusing depth features in the topo map setting, other variants do not take depths as model inputs.}\label{tab:hybrid}
\vspace{-6mm}
\end{table}

\vspace{1mm}
\nbf{2) The dependency on depth sensors}
We adopt in-domain pre-trained RedNet~\cite{jiang2018rednet} for depth estimation and then investigate BEVBert's dependence on depth sensors.
As shown in Tab.~\ref{tab:hybrid} (Row 3 \textit{v.s.} Row 4, Row 5 \textit{v.s.} Row 6), there is almost no performance drop when applying estimated depths for metric mapping. 
This suggests that our approach does not highly rely on accurate depth sensing. 
The main reason is that our metric maps are constructed in feature space, where we use rough grid depths (\eg, $14\times14$) for feature projection. 
We believe BEVBert has the potential to be extended in large-scale training with synthetic environments~\cite{koh2021pathdreamer,koh2022simple}, where depth sensors are unavailable.

\definecolor{Gray}{gray}{0.9}
\begin{table}[!htbp]
\vspace{-2mm}
\centering
\resizebox{0.45\textwidth}{!}{\begin{tabular}{cl | ccccc}
\toprule
\multicolumn{1}{c}{\#} & \multicolumn{1}{c}{Proxy Tasks} &
\multicolumn{1}{c}{TL} & \multicolumn{1}{c}{NE$\downarrow$} & \multicolumn{1}{c}{OSR$\uparrow$} & \multicolumn{1}{c}{\textbf{SR}$\uparrow$} & \multicolumn{1}{c}{\textbf{SPL}$\uparrow$} \\
\midrule
1 & None & 15.56 & 4.36 & 73.61 & 60.24 & 48.29 \\
2 & MLM & 16.26 & 3.09 & 83.82 & 73.52 & 60.13  \\
3 & MLM + HSAP & 14.50 & 3.03 & 82.67 & 74.03 & 63.03  \\
4 & MLM + HSAP + MSI & 14.55 & 2.81 & 83.65 & \color{blue}74.88 & \color{blue}63.60 \\
\bottomrule
\end{tabular}}
\vspace{-2.5mm}
\caption{Ablation study of pre-training tasks.}\label{tab:proxy}
\vspace{-3.5mm}
\end{table}
\nbf{3) The effect of pre-training tasks}
Tab.~\ref{tab:proxy} illustrates the effect of different pre-training tasks. 
Row 1 trains the model from scratch. It has the worst performance because the learned map lacks generic multimodal representations. 
With the generic MLM task, Row 2 can achieve decent performance (\eg, 73.52 \texttt{SR} and 60.13 \texttt{SPL}). However, the \texttt{TL} is high, thus leading to lower \texttt{SPL} compared to Row 3 and Row 4. 
In Row 3, the \texttt{TL} decreases, and \texttt{SPL} increases significantly after applying the HSAP task (\eg, $\uparrow$ 2.90 \texttt{SPL} over Row 2). 
It indicates that action prediction tasks are beneficial to learn action-informed map representations for efficient navigation.
Row 4 further improves the navigation performance with the proposed MSI task (\eg, $\uparrow$ 0.85 \texttt{SR} and $\uparrow$ 0.57 \texttt{SPL} over Row 3). 
The potential reason is that the agent learns to imagine unobserved areas and reduce the uncertainty for decision-making, which helps generalize unseen environments.

\definecolor{Gray}{gray}{0.9}
\begin{table}[!htbp]
\centering
\resizebox{0.48\textwidth}{!}{\begin{tabular}{clccc | cccc}
\toprule
 \multicolumn{1}{c}{\#} & \multicolumn{1}{c}{Scale} & \multicolumn{1}{c}{Cell Size} & \multicolumn{1}{c}{Map Size} & \multicolumn{1}{c}{Flops} & 
\multicolumn{1}{c}{NE$\downarrow$} & \multicolumn{1}{c}{OSR$\uparrow$} & \multicolumn{1}{c}{\textbf{SR}$\uparrow$} & \multicolumn{1}{c}{\textbf{SPL}$\uparrow$} \\
\midrule
1 & $11\times11$ & $0.5m^2$ & $5.5m^2$  & 4.5G   & 2.98 & 81.61 & 73.27 & 63.07 \\
2 & $11\times11$ & $1.0m^2$ & $11.0m^2$ & 4.5G   & 2.82 & 83.01 & 74.58 & 63.37 \\
3 & $21\times21$ & $0.5m^2$ & $10.5m^2$ & 15.2G  & 2.81 & 83.65 & \color{blue}74.88 & 63.60 \\
4 & $31\times31$ & $0.5m^2$ & $15.5m^2$ & 32.7G  & 2.83 & 83.23 & 74.84 & \color{blue}64.88 \\
\bottomrule
\end{tabular}}
\vspace{-2.5mm}
\caption{The effect of metric maps scale and size. Scales are set to odd to ensure the agent is at the central cell.}\label{tab:scale}
\vspace{-3.5mm}
\end{table}
\nbf{4) Scale and size of metric maps}
Tab.~\ref{tab:scale} reports BEVBert's performance using different scales and sizes of metric maps and the short-term transformer flops.
There is an upward trend in performance as the map size increases (Row 2 \textit{v.s.} Row 1), because the agent could perceive environments in a boarder scope.
Row 3 performs slightly better than Row 2 when the cell size decreases, which can be contributed to a better perception of minor objects. 
With a larger map scale, Row 4's performance does not increase obviously. 
The potential reason lies in the topo map used to capture long-range navigation dependency; thus, a large metric map only brings marginal benefit.
On the other hand, a larger metric map causes heavy computation (\eg, flops of the transformer are approximately quadratic w.r.t. the map scale). 
Therefore, Row 3 is our default setting.

\nbf{5) The effect of multi-step integration for metric maps}
We devise a local integration strategy for metric mapping in \S~\ref{sec:map}, which incorporates historical observations from visited nodes within $\kappa$ order. Tab.~\ref{tab:tmu} presents the effect of $\kappa$. 
With $\kappa=0$, the metric map is constructed from the current node's observations alone. It has the worst performance due to the lack of historical information, which may confuse the agent to understand mentioned short-term temporal dependency, such as ``keep the exhibit board on your right, go ...''.
When incorporating 1st-order historical observations, Row 2 improves \texttt{SPL} by 1.23 over Row 1, but no more gain as $\kappa$ goes up in Row 3. Because 1st-order integration is enough for a small local map.
\definecolor{Gray}{gray}{0.9}
\begin{table}[!htbp]
\vspace{-2.5mm}
\centering
\resizebox{0.3\textwidth}{!}{\begin{tabular}{cc | ccccc}
\toprule
\multicolumn{1}{c}{\#} & \multicolumn{1}{c}{$\kappa$} &
\multicolumn{1}{c}{TL} & \multicolumn{1}{c}{NE$\downarrow$} & \multicolumn{1}{c}{OSR$\uparrow$} & \multicolumn{1}{c}{\textbf{SR}$\uparrow$} & \multicolumn{1}{c}{\textbf{SPL}$\uparrow$} \\
\midrule
1 & 0 & 14.43 & 3.01 & 82.12 & 73.73 & 62.37 \\
2 & 1 & 14.55 & 2.81 & 83.65 & 74.88 & \color{blue}63.60 \\
3 & 2 & 14.89 & 2.81 & 84.29 & \color{blue}75.18 & 62.71 \\
\bottomrule
\end{tabular}}
\vspace{-2.5mm}
\caption{The effect of order $\kappa$ in metric mapping.}\label{tab:tmu}
\vspace{-4.5mm}
\end{table}

\vspace{1mm}
\nbf{6) Visual features}
BEVBert achieves better performance with  CLIP pre-trained features as shown in Tab.~\ref{tab:feature}.
Imagenet features may lack diverse visual concepts because they are learned by a one-hot classification task that focuses on salient regions of images. 
By contrast, CLIP features are learned by large-scale image-text matching, where visual grid features are informed by diverse linguistic concepts~\cite{shen2021much}, which can be more suitable for metric mapping. 
\definecolor{Gray}{gray}{0.9}
\begin{table}[!htbp]
\vspace{-3mm}
\centering
\resizebox{0.45\textwidth}{!}{\begin{tabular}{cl | ccccc}
\toprule
\multicolumn{1}{c}{\#} & \multicolumn{1}{c}{Features} &
\multicolumn{1}{c}{TL} & \multicolumn{1}{c}{NE$\downarrow$} & \multicolumn{1}{c}{OSR$\uparrow$} & \multicolumn{1}{c}{\textbf{SR}$\uparrow$} & \multicolumn{1}{c}{\textbf{SPL}$\uparrow$} \\
\midrule
1 & ViT-B/16-ImageNet~\cite{deng2009imagenet} & 15.90 & 2.91 & 83.44 & 74.03 & 61.86 \\
2 & ViT-B/16-CLIP~\cite{radford2021learning}  & 14.55 & 2.81 & 83.65 & \color{blue}74.88 & \color{blue}63.60 \\
\bottomrule
\end{tabular}}
\vspace{-2.5mm}
\caption{Comparison of different visual features.}\label{tab:feature}
\vspace{-4mm}
\end{table}

%% file: 10_conclusion.tex
\vspace{-2mm}
\section{Conclusion}
\label{sec:conclusion}
In this paper, we first devise a hybrid map to balance the demand of VLN for both short-term reasoning and long-term planning.
Based on the hybrid map, we propose a new pre-training paradigm, BEVBert, to learn visual-textual associations in an explicit spatial representation.
We empirically validate that the learned multimodal map representations could enhance spatial-aware cross-modal reasoning and facilitate the final language-guided navigation goal.  
Extensive experiments demonstrate the effectiveness of the proposed method and BEVBert achieves state-of-the-art.

\section{Acknowledgments}
This work was partly supported by National Key Research and Development Program of China Grant No. 2018AAA0100400, National Natural Science Foundation of China (62236010 and 62276261), and Key Research Program of Frontier Sciences CAS Grant No. ZDBS-LYJSC032. 
We warmly thank ICCV reviewers, Enze Xie, Yicong Hong, and Zun Wang for their valuable suggestions that have helped improve the soundness and quality of this paper.

%% file: 12_appendix.tex
\appendix
\label{sec:appendix}
{\large
\noindent\textbf{Appendices}
}
\vspace{1mm}

Section~\ref{sec:sup_dataset} presents more details about evaluation datasets and metrics. 
Model variants and training objectives are described in Section~\ref{sec:sup_model}.
Details about experimental setups are provided in Section~\ref{sec:sup_exprs}, and more comparisons against state-of-the-art methods are shown in Section~\ref{sec:sup_sota}.
Finally, we present several visualizations of failure cases in Section~\ref{sec:sup_viz}.

\begin{table*}[h]
\centering
\small
\resizebox{0.98\textwidth}{!}{\begin{tabular}{cccccccccccc} \toprule
\multirow{2}{*}{Task Type} & \multirow{2}{*}{Granularity} & \multirow{2}{*}{Dataset} & \multicolumn{2}{c}{Train} & \multicolumn{2}{c}{Val Seen} & \multicolumn{2}{c}{Val Unseen} & \multicolumn{2}{c}{Test Unseen} \\
& & & \#house & \#instr & \#house & \#instr & \#house & \#instr & \#house & \#instr \\ \midrule
\multirow{3}{*}{Instruction-following} & \multirow{3}{*}{Fine-grained} 
  & R2R~\cite{anderson2018vision}   & 61 & 14,039 & 56 & 1,021 & 11 & 2,349  & 18 & 4,173 \\
& & R2R-CE~\cite{krantz2020beyond}  & 61 & 10,819 & 53 & 778   & 11 & 1,839  & 18 & 3,408 \\
& & RxR~\cite{ku2020room}           & 60 & 79,467 & 58 & 8,813 & 11 & 13,652 & 17 & 12,249 \\ 
\midrule
Goal-oriented & Coarse-grained & REVERIE~\cite{qi2020reverie} & 60 & 10,466 & 46 & 1,423 & 10 & 3,521 & 16 & 6,292 \\ \bottomrule
\end{tabular}}
\vspace{-2.5mm}
\caption{Dataset statistics. \#house, \#instr denote the number of houses and instructions respectively.}\label{tab:dataset_stats}
\end{table*}

\section{Evaluation Datasets and  Metrics}\label{sec:sup_dataset}
Our approach is evaluated on R2R~\cite{anderson2018vision}, R2R-CE~\cite{krantz2020beyond}, RxR~\cite{ku2020room} and REVERIE~\cite{qi2020reverie} datasets, which are built upon the Matterport3D~\cite{chang2017matterport3d} indoor scene dataset. 
We summarize the dataset statistics in Tab.~\ref{tab:dataset_stats}. The four datasets differ in task type (instruction-following \textit{v.s.} goal-oriented) and instructions granularity (fine-grained \textit{v.s.} coarse-grained).

\vspace{1mm}
\noindent\textbf{Room-to-Room (R2R)}~\cite{anderson2018vision} provides fine-grained (step-by-step) instructions. 
The agent is required to follow an instruction to reach the target location.
It receives panoramic observations at each viewpoint, and navigation is simplified as transporting among viewpoints on the predefined graph of an environment. 
The dataset contains 61 houses for training, 56 houses for validation in seen environments, 11 and 18 houses for validation and testing in unseen environments, respectively. 
Each instruction in R2R describes a full path, such as ``\textit{Head straight until you pass the wall with holes in it the turn left and wait by the glass table with the white chairs.}''.

\vspace{1mm}
\noindent\textbf{Room-to-Room in Continuous Environments (R2R-CE)}~\cite{krantz2020beyond} extends R2R in continuous environments.
It discards the predefined graph assumption, instead, requiring the agent to navigate freely with low-level actions (\eg, \texttt{FORWARD} 0.25m, \texttt{ROTATE} 15\degree) on 3D meshes~\cite{savva2019habitat}. 
R2R-CE contains a subset of R2R's instruction-path pairs because some paths are unusable on the 3D meshes.

\vspace{1mm}
\noindent\textbf{Room-across-Room (RxR)}~\cite{ku2020room} is a more challenging instruction-following dataset, which emphasizes the role of language guidance and provides large-scale multilingual instructions (en-IN, en-US, hi-IN, te-IN). 
Instructions in RxR involve more descriptions of visual entities and relations, such as ``\textit{You are facing towards a wall, turn around and move forward. You are now standing in front of a glass cabin and on your right side you have a table with a computer. You can see a table with a black chair right in front of you. Walk between these two tables and move forward. Now you can see a table with orange chair on your right side. Move forward towards the centre table which is right in front of a couch and that is your end point.}''.
Besides, annotated paths in RxR are much longer than R2R ($\sim 2$ times) and the ground truth paths are not the shortest path between the starting and ending points.

\vspace{1mm}
\noindent\textbf{REVERIE}~\cite{qi2020reverie} is a goal-oriented dataset which provides coarse-grained instructions. 
Instructions in REVERIE are concise and mainly describe target rooms and target objects, such as ``\textit{Go to the office and clean the black and white picture of a child}''. 
The task focuses more on knowledge and commonsense exploitation rather than instruction following. 
Furthermore, the agent must identify the target object from a set of candidates after reaching the desired location.
The dataset has 4,140 target objects in 489 categories, and each target viewpoint has 7 objects on average.

\vspace{1mm}
\noindent\textbf{Evaluation Metrics}. VLN tasks mainly focus on the agent's generalization ability in unseen environments (val unseen and test unseen splits).
The main evaluation metrics of the above datasets are slightly different.
On the R2R/R2R-CE dataset, \texttt{SR} and \texttt{SPL} are the main metrics to evaluate the navigation accuracy and efficiency, where a predicted path is regarded as \textit{success} if the agent stops within 3 meters of the target location. 
RxR dataset does not have shortest-path prior, and it additionally takes \texttt{NDTW} and \texttt{SDTW} to measure the fidelity between predicted and annotated paths. The two metrics reflect how well the agent interprets and follows instructions. 
On the REVERIE dataset, annotated paths have shortest-path prior and the main navigation metrics are \texttt{SR} and \texttt{SPL}. A predicted path is considered as \textit{success} if the target object is visible within 3 meters at the endpoint. 
It additionally uses \texttt{RGS} and \texttt{RGSPL} to evaluate the object grounding capacity of the agent, an grounding is \textit{success} if the predicted and annotated objects are the same.

\section{Model Details}\label{sec:sup_model}
\begin{figure*}[h]
\centering
\includegraphics[width=0.88\textwidth]{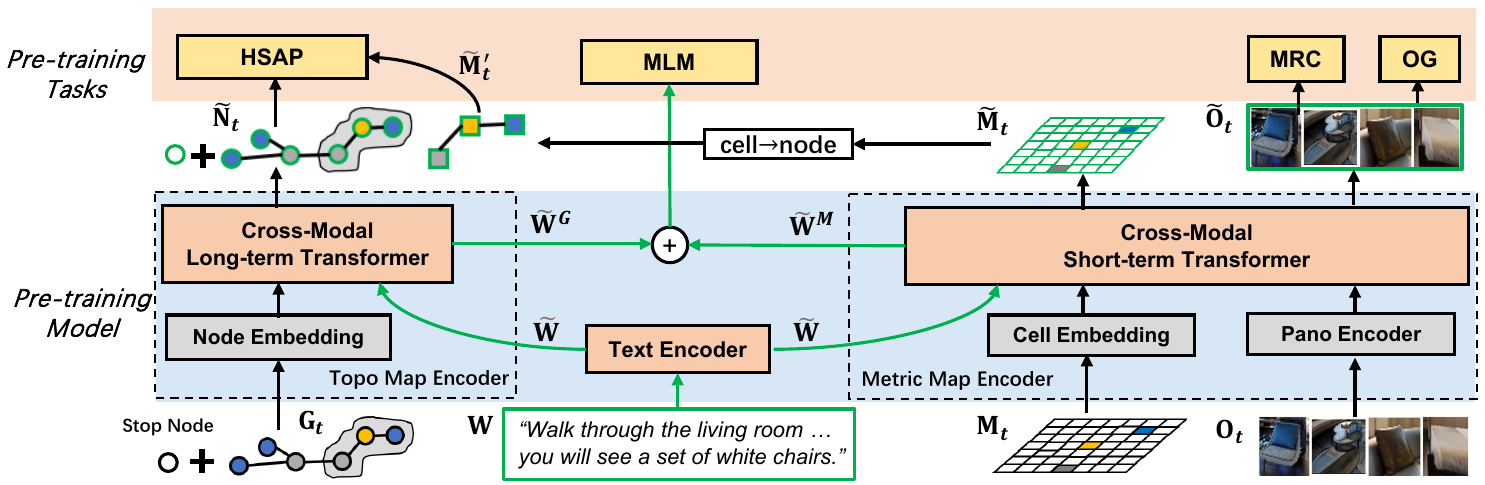}
\vspace{-3mm}
\caption{
Adapt the proposed pre-training model to the REVERIE task. 
Additional object features $\mathbf{O}_t$ are fed into the metric map encoder to obtain multimodal object representations $\widetilde{\mathbf{O}}_{t}$. 
We use MRC and OG tasks to learn cross-modal object reasoning and grounding.
}\label{fig:arch_reverie}
\vspace{-3mm}
\end{figure*}

\subsection{Adaptation to the REVERIE Dataset}
REVERIE provides candidate object annotations at each step and requires the agent to point out the target object when it stops.
As shown in Fig.~\ref{fig:arch_reverie}, we feed these candidates into the short-term branch (metric map encoder) to enable object grounding. 
Specifically, at each step, we first use the same ViT as in Section~\ref{sec:map} to extract object features $\mathbf{O}_t=\{ \mathbf{o}_z | \mathbf{o}_z \in \mathcal{R}^D \}_{z=1}^Z$.
After adding position embeddings (sine and cosine of orientations~\cite{hong2021vln,chen2022think,lin2021scene}), these object features are concatenated with view feature vectors $\mathbf{V}_{t}^{p}$ and fed into the pano encoder in Section~\ref{sec:map} to obtain contextual object embeddings. 
Then, cell and object embeddings are concatenated as the visual modality while encoded instructions as the linguistic modality.
We feed them into the short-term transformer to perform cross-modal reasoning as explained in Section~\ref{sec:metric_encoder}.
The output multimodal object representations $\widetilde{\mathbf{O}}_t = \{ \tilde{o}_z \}_{z=1}^{Z}$ are learned via MRC and OG tasks, which will be detailed in the next section.

\subsection{Pre-training Objectives}\label{sec:pretrain_reverie}
\noindent\textbf{R2R/R2R-CE and RxR.} We sample pre-training tasks for each mini-batch to train the BEVBert model. The sampling ratio for R2R/R2R-CE and RxR datasets is $\textrm{MLM}:\textrm{HSAP}:\textrm{MSI}=5:5:1$.
We randomly chunk a sampled expert trajectory $\mathbf{\Gamma}$ from head to obtain $\mathbf{\Gamma'}$ for offline map construction.

\vspace{1mm}
\noindent\textbf{REVERIE.} Instructions in REVERIE mainly describe the rooms and target objects at endpoints. 
We do not employ MSI task due to the lack of intermediate path descriptions; instead, Masked Region Classification (MRC)~\cite{lu2019vilbert} and Object Grounding (OG)~\cite{lin2021scene} are used for final object reasoning and grounding. 
MRC aims to predict semantic labels of masked objects by reasoning over the surrounding objects and the instructions. We randomly mask objects with a 15\% probability and feed them into the metric map encoder.
The semantic labels of objects are class probability predicted by a ViT pre-trained on ImageNet~\cite{deng2009imagenet}. The task is optimized by minimizing the KL divergence between the predicted and target probability distribution. 
OG is a downstream-specific task. After obtaining the multimodal object representation $\widetilde{\mathbf{O}}_t$, a two-layer feed-forward network is employed to predict the target object scores. Given the target object label $\mathbf{o}^{*}$ at step $T$ (the endpoint), the task is optimized by minimizing the negative log-likelihood:
{\small
\begin{equation}\label{eq:og_loss}
\mathcal{L}_{\textrm{OG}} = -\log \mathcal{P}_{\theta} (\mathbf{o}^{*} | \mathbf{W}, \mathbf{\Gamma}, \mathbf{O}_{T})   
\end{equation}
}%
We sample pre-training tasks for each mini-batch to train the model on REVERIE dataset, and the sampling ratio is $\textrm{MLM}:\textrm{HSAP}:\textrm{MRC}:\textrm{OG}=1:1:1:1$.

\subsection{Layer Variants in Fine-tuning}
During fine-tuning, the computational graph gradually expands as the trajectory rolls out and may cause GPU out of memory in extreme cases.  
To alleviate the problem, for each transformer layer of the two map encoders, we cut off self-attention and cross-attention of the text branch, \ie, the language-to-vision cross-attention and the language-to-language self-attention. 
All transformer layers share the same text representations. We empirically found it does not severely hamper navigation performance, and the same phenomenon is also observed in RecBert~\cite{hong2021vln} and HAMT~\cite{chen2021history}.

\begin{figure*}[h]
\vspace{-3mm}
\centering
\includegraphics[width=0.90\textwidth]{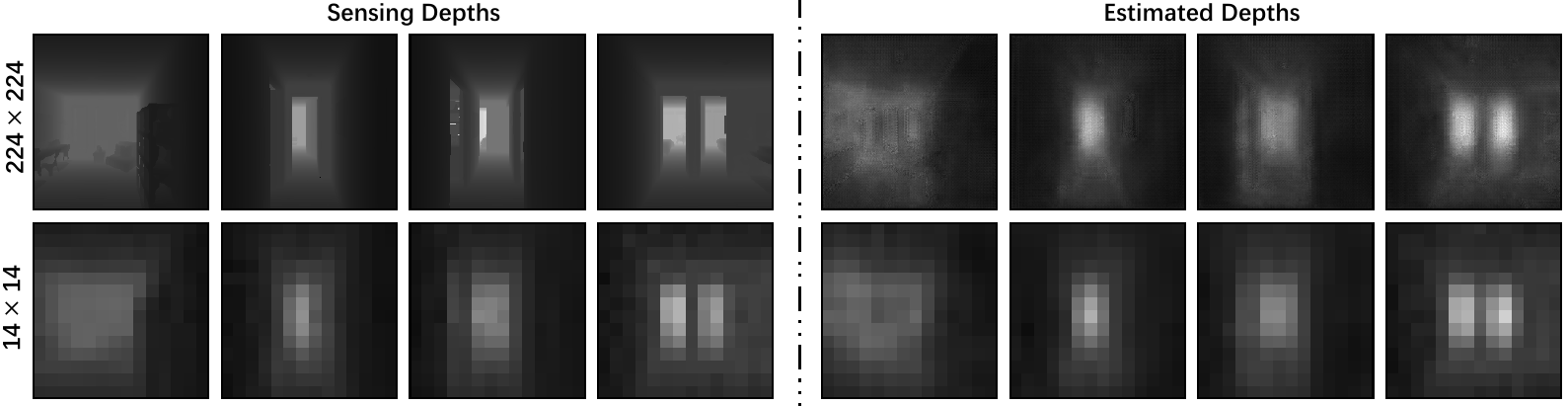}
\vspace{-3mm}
\caption{
Qualitative visualization of the sensing and estimated depths. The top row represents depths of the original scale, and the bottom row presents downsized depths that have the same scale with grid features.
}\label{fig:depth}
\vspace{-3mm}
\end{figure*}

\subsection{Fine-tuning Objectives}
During fine-tuning, we alternatively run `teacher-forcing' $\mathcal{L}_{\mathrm{TF}}$ and `student-forcing' $\mathcal{L}_{\mathrm{SF}}$. The model is optimized by the mixed loss $\mathcal{L}$, formally:
{\small
\begin{equation}\label{eq:ts_r2r}
\begin{aligned}
\mathcal{L}_{\textrm{TF}} &= -\sum\nolimits_{t=1}^T \log \mathcal{P}_{\theta} (\mathbf{a}_{t}^{*} | \mathbf{W}, \mathbf{\Gamma}_{< t}) \\
\mathcal{L}_{\textrm{SF}} &= -\sum\nolimits_{t=1}^T \log \mathcal{P}_{\theta} (\mathbf{a}_{t}^{G*} | \mathbf{W}, \widetilde{\mathbf{\Gamma}}_{< t}) \\
\mathcal{L} &= \mathbf{\lambda} \cdot \mathcal{L}_{\textrm{TF}} + \mathcal{L}_{\textrm{SF}}
\end{aligned}
\end{equation}
}%

In $\mathcal{L}_{\textrm{TF}}$, the agent executes ground-truth actions $\mathbf{a}_{t}^{*}$ to follow the expert trajectory $\mathbf{\Gamma}$, and the loss is the accumulation of negative log-likelihoods of these expert actions.
In $\mathcal{L}_{\textrm{SF}}$, the agent generates trajectory $\tilde{\mathbf{\Gamma}}$ by on-policy action sampling and supervised by pseudo labels $\mathbf{a}_{t}^{G*}$.
Specifically, the action is sampled via the predicted action probability distribution at each step, and the pseudo labels are determined via goal-oriented or fidelity-oriented heuristics. 
A goal-oriented label is a ghost node that has the shortest path length to the final target location, while a fidelity-oriented label is a ghost node through which the sampled path has the highest fidelity (\texttt{NDTW}) with the expert path.
We use goal-oriented labels for R2R/R2R-CE and REVERIE, while fidelity-oriented labels for RxR because it does not have shortest-path prior.
Besides, OG loss is added for fine-tuning on REVERIE:
{\small
\begin{equation}\label{eq:ts_reverie}
\mathcal{L} = \mathbf{\lambda} \cdot \mathcal{L}_{\textrm{TF}} + \mathcal{L}_{\textrm{SF}} + \mathcal{L}_{\textrm{OG}}
\end{equation}
}%

The `teacher-forcing' encourages the agent to follow the expert path, while the `student-forcing' encourages the agent to explore the environment. 
We set the balanced term $\mathbf{\lambda}=0.2$ on R2R/R2R-CE and REVERIE datasets, while $\mathbf{\lambda}=0.8$ on the RxR dataset. Because annotated paths in RxR are much longer, a small $\mathbf{\lambda}$ can lead to unnecessary exploration and hamper navigation fidelity.

\section{Experimental Setups}\label{sec:sup_exprs}

%

\subsection{Depth Estimation}
This section details the depth ablation study in Section~\ref{sec:ablation}. 
We employ RedNet~\cite{jiang2018rednet} for depth estimation, which takes RGB images as inputs and uses a U-Net-like architecture for depth regression.
The estimated depths are outputs of a sigmoid layer, and downsized depth images also supervise intermediate layers to speed up convergence. 
We train the model in train-split houses of Matterport3D dataset~\cite{chang2017matterport3d}.
Then, a trained model is used to estimate depths for all viewpoints of all houses, and we retrain BEVBert with these pseudo depths.
We visualize some sensing and estimated depths in Fig.~\ref{fig:depth}.
Intuitively, estimated depths are of low quality and have noise, but their quality is similar to sensing depths' after downsampling.
This can also explain why BEVBert does not rely on very accurate depth images.

\subsection{Spatial and Numerical Tokens}
Tab.~\ref{tab:token} summarizes the token templates we used to extract spatial and numerical related instructions in Section~\ref{sec:analysis}.
Extracted instructions are grouped via the number of special tokens in each instruction.
To ensure reliable performance estimation, we omit those groups which contain less than 40 instructions.

\begin{table}[h]
\begin{adjustbox}{width=\linewidth,center}
\begin{tabular}{l c }
\cmidrule[\heavyrulewidth]{1-2}
\textbf{Token Type} & \textbf{Token Templates} \\ \cmidrule[\heavyrulewidth]{1-2}
Spatial     & \makecell[l]{
on the left, on your left, to the left, to your left \\
left of, left side of, leftmost, on the right, \\
on your right, to the right, to your right, right of, \\
right side of, rightmost, near, nearest, behind, \\
between, next to, end of, edge of, front of, \\
middle of, top of, bottom of
}
\\ \cmidrule{1-2}
Numerical     &  \makecell[l]{
first, second, third, fourth, fifth, sixth, seventh, \\
eighth, one, two, three, four, five, six, seven, \\
eight, 1, 2, 3, 4, 5, 6, 7, 8
}
\\ \cmidrule{1-2}
\end{tabular}
\end{adjustbox}
\vspace{-2.5mm}
\caption{Templates of spatial and numerical tokens.}
\label{tab:token}
\vspace{-2.5mm}
\end{table}

\definecolor{Gray}{gray}{0.9}
\begin{table*}[h]
\centering
\resizebox{0.92\textwidth}{!}{\begin{tabular}{lccccc | ccccc | ccccc}
\toprule 
\multicolumn{1}{c}{} & \multicolumn{5}{c}{Val Seen} & \multicolumn{5}{c}{Val Unseen} & \multicolumn{5}{c}{Test Unseen} \\
\cmidrule(r){2-6} \cmidrule(r){7-11} \cmidrule(r){12-16}
\multicolumn{1}{c}{Methods} & 
\multicolumn{1}{c}{TL} & \multicolumn{1}{c}{NE$\downarrow$} & \multicolumn{1}{c}{OSR$\uparrow$} & \multicolumn{1}{c}{\textbf{SR}$\uparrow$} & \multicolumn{1}{c}{\textbf{SPL}$\uparrow$} & 
\multicolumn{1}{c}{TL} & \multicolumn{1}{c}{NE$\downarrow$} & \multicolumn{1}{c}{OSR$\uparrow$} & \multicolumn{1}{c}{\textbf{SR}$\uparrow$} & \multicolumn{1}{c}{\textbf{SPL}$\uparrow$} & 
\multicolumn{1}{c}{TL} & \multicolumn{1}{c}{NE$\downarrow$} & \multicolumn{1}{c}{OSR$\uparrow$} & \multicolumn{1}{c}{\textbf{SR}$\uparrow$} & \multicolumn{1}{c}{\textbf{SPL}$\uparrow$} \\
\midrule
Human                   
& - & - & - & - & - 
& - & - & - & - & - 
& 11.85 & 1.61 & 90 & 86 & 76 \\
\midrule
Seq2Seq~\cite{anderson2018vision}
& 11.33 & 6.01 & 53 & 39 & - 
& 8.39 & 7.81 & 28 & 21 & - 
& 8.13 & 7.85 & 27 & 20 & - \\
SF~\cite{fried2018speaker}
& - & 3.36 & 74 & 66 & - 
& - & 6.62 & 45 & 36 & - 
& 14.82 & 6.62 & - & 35 & 28  \\
Chasing~\cite{anderson2019chasing}
& 10.15 & 7.59 & 42 & 34 & 30
& 9.64 & 7.20 & 44 & 35 & 31
& 10.03 & 7.83 & 42 & 33 & 30 \\
RCM~\cite{wang2019reinforced}
& 10.65 & 3.53 & 75 & 67 & - 
& 11.46 & 6.09 & 50 & 43 & - 
& 11.97 & 6.12 & 50 & 43 & 38 \\
SM~\cite{ma2019self}
& - & 3.22 & 78 & 67 & 58
& - & 5.52 & 56 & 45 & 32 
& 18.04 & 5.67 & 59 & 48 & 35 \\
EnvDrop~\cite{tan2019learning}
& 11.00 & 3.99 & - & 62 & 59
& 10.70 & 5.22 & - & 52 & 48
& 11.66 & 5.23 & 59 & 51 & 47 \\
OAAM~\cite{qi2020object}
& - & - & 73 & 65 & 62
& - & - & 61 & 54 & 50
& - & - & 61 & 53 & 50 \\
AuxRN~\cite{zhu2020vision}
& - & 3.33 & 78 & 70 & 67
& - & 5.28 & 62 & 55 & 50
& - & 5.15 & 62 & 55 & 51 \\
Active~\cite{wang2020active}
& - & 3.20 & 80 & 70 & 52
& - & 4.36 & 70 & 58 & 40
& - & 4.33 & 71 & 60 & 41 \\
NvEM~\cite{an2021neighbor}
& 11.09 & 3.44 & - & 69 & 65
& 11.83 & 4.27 & - & 60 & 55
& 12.98 & 4.37 & 66 & 58 & 54 \\
SEvol~\cite{chen2022reinforced}
& 11.97 & 3.56 & - & 67 & 63
& 12.26 & 3.99 & - & 62 & 57
& 13.40 & 4.13 & - & 62 & 57 \\
SSM~\cite{wang2021structured}
& 14.70 & 3.10 & 80 & 71 & 62
& 20.70 & 4.32 & 73 & 62 & 45
& 20.40 & 4.57 & 70 & 61 & 46 \\
PREVAL~\cite{hao2020towards}\dag
& 10.32 & 3.67 & - & 69 & 65
& 10.19 & 4.71 & - & 58 & 53
& 10.51 & 5.30 & 61 & 54 & 51 \\
AirBert~\cite{guhur2021airbert}\dag
& 11.09 & 2.68 & - & 75 & 70
& 11.78 & 4.10 & - & 62 & 56 
& 12.41 & 4.13 & - & 62 & 57 \\
RecBert~\cite{hong2021vln}\dag
& 11.13 & 2.90 & - & 72 & 68
& 12.01 & 3.93 & - & 63 & 57
& 12.35 & 4.09 & 70 & 63 & 57 \\
REM~\cite{liu2021vision}\dag
& 10.88 & 2.48 & - & 75 & 72
& 12.44 & 3.89 & - & 64 & 58
& 13.11 & 3.87 & 72 & 65 & 59 \\
HAMT~\cite{chen2021history}\dag
& 11.15 & 2.51 & - & 76 & 72
& 11.46 & 3.65 & - & 66 & 61
& 12.27 & 3.93 & 72 & 65 & 60 \\
EnvEdit*~\cite{li2022envedit}\dag
& 11.18 & 2.32 & - & 77 & \color{blue}74
& 11.13 & 3.24 & - & 69 & \color{blue}64 
& 11.90 & 3.59 & - & 68 & \color{blue}64 \\
TD-STP~\cite{zhao2022target}\dag
& - & 2.34 & 83 & 77 & 73 
& - & 3.22 & 76 & 70 & 63 
& - & 3.73 & 72 & 67 & 61 \\
DUET~\cite{chen2022think}\dag
& 12.32 & 2.28 & 86 & 79 & 73 
& 13.94 & 3.31 & 81 & 72 & 60
& 14.73 & 3.65 & 76 & 69 & 59 \\
BEVBert (Ours)\dag
& 13.56 & \color{blue}2.17 & \color{blue}88 & \color{blue}81 & \color{blue}74
& 14.55 & \color{blue}2.81 & \color{blue}84 & \color{blue}75 & \color{blue}64
& 15.87 & \color{blue}3.13 & \color{blue}81 & \color{blue}73 & \color{blue}62 \\
\bottomrule
\end{tabular}}
\vspace{-2.5mm}
\caption{Comparison with state-of-the-art methods on R2R dataset. *Ensemble of three agents. \dag~denotes pre-training-based methods.}\label{tab:r2r_full}
\vspace{-2.5mm}
\end{table*}

\definecolor{Gray}{gray}{0.9}
\begin{table*}[h]
\centering
\resizebox{0.88\textwidth}{!}{\begin{tabular}{lcccc | cccc | cccc}
\toprule 
\multicolumn{1}{c}{} & \multicolumn{4}{c}{Val Seen} & \multicolumn{4}{c}{Val Unseen} & \multicolumn{4}{c}{Test Unseen} \\
\cmidrule(r){2-5} \cmidrule(r){6-9} \cmidrule(r){10-13}
\multicolumn{1}{c}{Methods} & 
\multicolumn{1}{c}{NE$\downarrow$} & \multicolumn{1}{c}{\textbf{SR}$\uparrow$} & \multicolumn{1}{c}{\textbf{NDTW}$\uparrow$} & \multicolumn{1}{c}{\textbf{SDTW}$\uparrow$} &
\multicolumn{1}{c}{NE$\downarrow$} & \multicolumn{1}{c}{\textbf{SR}$\uparrow$} & \multicolumn{1}{c}{\textbf{NDTW}$\uparrow$} & \multicolumn{1}{c}{\textbf{SDTW}$\uparrow$} &
\multicolumn{1}{c}{NE$\downarrow$} & \multicolumn{1}{c}{\textbf{SR}$\uparrow$} & \multicolumn{1}{c}{\textbf{NDTW}$\uparrow$} & \multicolumn{1}{c}{\textbf{SDTW}$\uparrow$} \\
\midrule
Human                   
& - & - & - & - 
& - & - & - & - 
& 0.9 & 93.9 & 79.5 & 76.9  \\
\midrule
LSTM~\cite{ku2020room}
& 10.7 & 25.2 & 42.2 & 20.7 
& 10.9 & 22.8 & 38.9 & 18.2
& 12.0 & 21.0 & 36.8 & 16.9 \\
EnvDrop+~\cite{shen2021much}
& - & - & - & -
& - & 42.6 & 55.7 & - 
& - & 38.3 & 51.1 & 32.4 \\
CLEAR-C~\cite{li2022clear}
& - & - & - & -
& - & - & - & -
& - & 40.3 & 53.7 & 34.9 \\
HAMT~\cite{chen2021history}
& - & 59.4 & 65.3 & 50.9
& - & 56.5 & 63.1 & 48.3
& 6.2 & 53.1 & 59.9 & 45.2 \\
EnvEdit*~\cite{li2022envedit}
& - & 67.2 & 71.1 & 58.5
& - & 62.8 & 68.5 & 54.6 
& 5.1 & 60.4 & 64.6 & 51.8 \\
BEVBert (Ours)
& 3.8 & 68.9 & 70.0 & 58.4
& 4.6 & 64.1 & 63.9 & 52.6
& - & - & - & - \\
BEVBert (Ours)
& \color{blue}3.2 & \color{blue}75.0 & \color{blue}76.3 & \color{blue}66.7
& \color{blue}4.0 & \color{blue}68.5 & \color{blue}69.6 & \color{blue}58.6
& \color{blue}4.8 & \color{blue}64.4 & \color{blue}65.4 & \color{blue}54.2 \\
\bottomrule
\multicolumn{12}{l}{\small{*Results from an ensemble of three agents.  Results without Marky synthetic instructions~\cite{wang2022less}.}}
\end{tabular}}
\vspace{-2.5mm}
\caption{Comparison with state-of-the-art methods on RxR dataset.}\label{tab:rxr_full}
\vspace{-2.5mm}
\end{table*}

\definecolor{Gray}{gray}{0.9}
\begin{table*}[h]
\centering
\resizebox{0.98\textwidth}{!}{\begin{tabular}{l ccccc | ccccc | ccccc }
\toprule
\multirow{3}{*}{Methods} & \multicolumn{5}{c}{Val seen} & \multicolumn{5}{c}{Val Unseen} & \multicolumn{5}{c}{Test Unseen} \\
\cmidrule(r){2-6} \cmidrule(r){7-11} \cmidrule(r){12-16} 
& \multicolumn{3}{c}{Navigation} & \multicolumn{2}{c}{Grounding} & \multicolumn{3}{c}{Navigation} & \multicolumn{2}{c}{Grounding} & \multicolumn{3}{c}{Navigation} & \multicolumn{2}{c}{Grounding} \\
& \multicolumn{1}{c}{OSR$\uparrow$} & \multicolumn{1}{c}{\textbf{SR}$\uparrow$} & \multicolumn{1}{c}{\textbf{SPL}} & \multicolumn{1}{c}{\textbf{RGS}$\uparrow$} & \multicolumn{1}{c}{\textbf{RGSPL}$\uparrow$} 
& \multicolumn{1}{c}{OSR$\uparrow$} & \multicolumn{1}{c}{\textbf{SR}$\uparrow$} & \multicolumn{1}{c}{\textbf{SPL}$\uparrow$} & \multicolumn{1}{c}{\textbf{RGS}$\uparrow$} & \multicolumn{1}{c}{\textbf{RGSPL}$\uparrow$} 
& \multicolumn{1}{c}{OSR$\uparrow$} & \multicolumn{1}{c}{\textbf{SR}$\uparrow$} & \multicolumn{1}{c}{\textbf{SPL}$\uparrow$} & \multicolumn{1}{c}{\textbf{RGS}$\uparrow$} & \multicolumn{1}{c}{\textbf{RGSPL}$\uparrow$} \\ 
\midrule
Human
& - & - & - & - & -
& - & - & - & - & -
& 81.51 & 53.66 & 86.83 & 77.84 & 51.44 \\
\midrule
Seq2Seq~\cite{anderson2018vision} 
& 35.70 & 29.59 & 24.01 & 18.97 & 14.96
& 8.07 & 4.20 & 2.84 & 2.16 & 1.63 
& 6.88 & 3.99 & 3.09 & 2.00 & 1.58 \\
RCM~\cite{wang2019reinforced} 
& 29.44 & 23.33 & 21.82 & 16.23 & 15.36
& 14.23 & 9.29 & 6.97 & 4.89 & 3.89 
& 11.68 & 7.84 & 6.67 & 3.67 & 3.14 \\
SMNA~\cite{ma2019self} 
& 43.29 & 41.25 & 39.61 & 30.07 & 28.98
& 11.28 & 8.15 & 6.44 & 4.54 & 3.61 
& 8.39 & 5.80 & 4.53 & 3.10 & 2.39 \\
FAST-MATTN~\cite{qi2020reverie} 
& 55.17 & 50.53 & 45.50 & 31.97 & 29.66
& 28.20 & 14.40 & 7.19 & 7.84 & 4.67 
& 30.63 & 19.88 & 11.6 & 11.28 & 6.08 \\
SIA~\cite{lin2021scene }
& 65.85 & 61.91 & 57.08 & 45.96 & 42.65
& 44.67 & 31.53 & 16.28 & 22.41 & 11.56 
& 44.56 & 30.80 & 14.85 & 19.02 & 9.20 \\
RecBERT~\cite{hong2021vln} 
& 53.90 & 51.79 & 47.96 & 38.23 & 35.61
& 35.20 & 30.67 & 24.90 & 18.77 & 15.27 
& 32.91 & 29.61 & 23.99 & 16.50 & 13.51 \\
AirBert~\cite{guhur2021airbert}
& 48.98 & 47.01 & 42.34 & 32.75 & 30.01
& 34.51 & 27.89 & 21.88 & 18.23 & 14.18 
& 34.20 & 30.26 & 23.61 & 16.83 & 13.28 \\
HAMT~\cite{chen2021history} 
& 47.65 & 43.29 & 40.19 & 27.20 & 25.18
& 36.84 & 32.95 & 30.20 & 18.92 & 17.28 
& 33.41 & 30.40 & 26.67 & 14.88 & 13.08 \\ 
TD-STP~\cite{zhao2022target} 
& - & - & - & - & -
& 39.48 & 34.88 & 27.32 & 21.16 & 16.56 
& 40.26 & 35.89 & 27.51 & 19.88 & 15.40 \\ 
DUET~\cite{chen2022think} 
& 73.86 & 71.75 & 63.94 & 57.41 & 51.14
& 51.07 & 46.98 & 33.73 & 32.15 & 23.03 
& 56.91 & 52.51 & 36.06 & 31.88 & 22.06 \\
BEVBert (Ours) 
& \color{blue}76.18 & \color{blue}73.72 & \color{blue}65.32 & \color{blue}57.70 & \color{blue}51.73
& \color{blue}56.40 & \color{blue}51.78 & \color{blue}36.37 & \color{blue}34.71 & \color{blue}24.44 
& \color{blue}57.26 & \color{blue}52.81 & \color{blue}36.41 & \color{blue}32.06 & \color{blue}22.09 \\
\bottomrule
\end{tabular}}
\vspace{-2.5mm}
\caption{Comparison with state-of-the-art methods on REVERIE dataset.}\label{tab:reverie_full}
\end{table*}
\section{More Comparisons with State-of-the-Art}\label{sec:sup_sota}
In Tab.~\ref{tab:r2r_full}, Tab.~\ref{tab:rxr_full} and Tab.~\ref{tab:reverie_full}, we present more comparisons with state-of-the-art methods on R2R, RxR and REVERIE, respectively. The main metrics of each dataset are highlighted. 
BEVBert also achieves state-of-the-art performance in seen splits on all metrics, but the performance is still far behind humans'.
For example, on the test unseen split of RxR dataset, humans can achieve 93.9 \texttt{SR} and 76.9 \texttt{SDTW}, while BEVBert has 64.4 \texttt{SR} and 54.2 \texttt{SDTW}.

\section{More Qualitative Examples}\label{sec:sup_viz}
\begin{figure*}[h]
\centering
\includegraphics[width=0.81\textwidth]{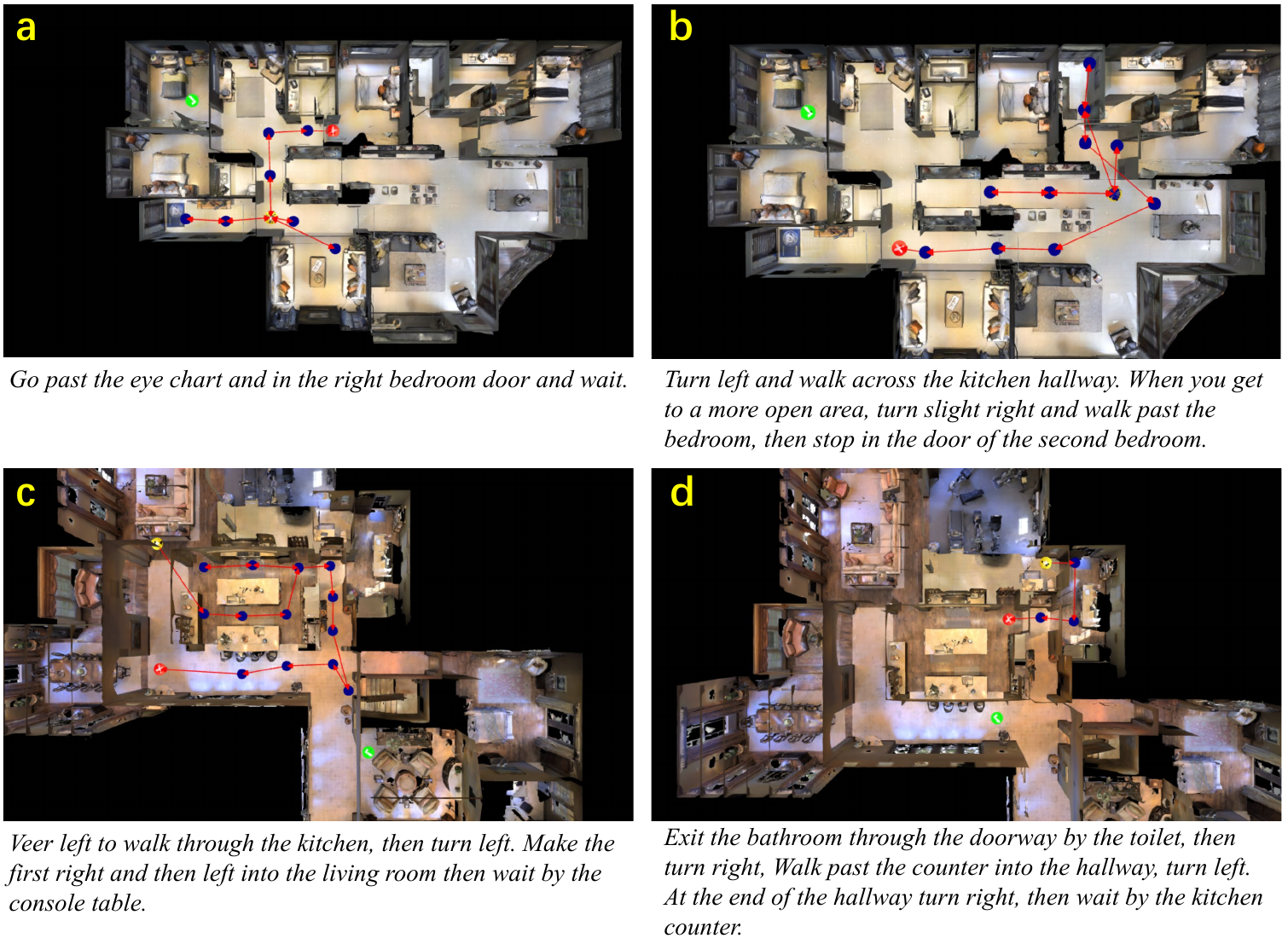}
\vspace{-3mm}
\caption{
Failure cases of `early lost' in val unseen splits of R2R. Yellow and green circles denote the start and target locations, respectively, and the red circles represent incorrect endpoints. 
}\label{fig:early}
\end{figure*}

\begin{figure*}[h]
\centering
\includegraphics[width=0.81\textwidth]{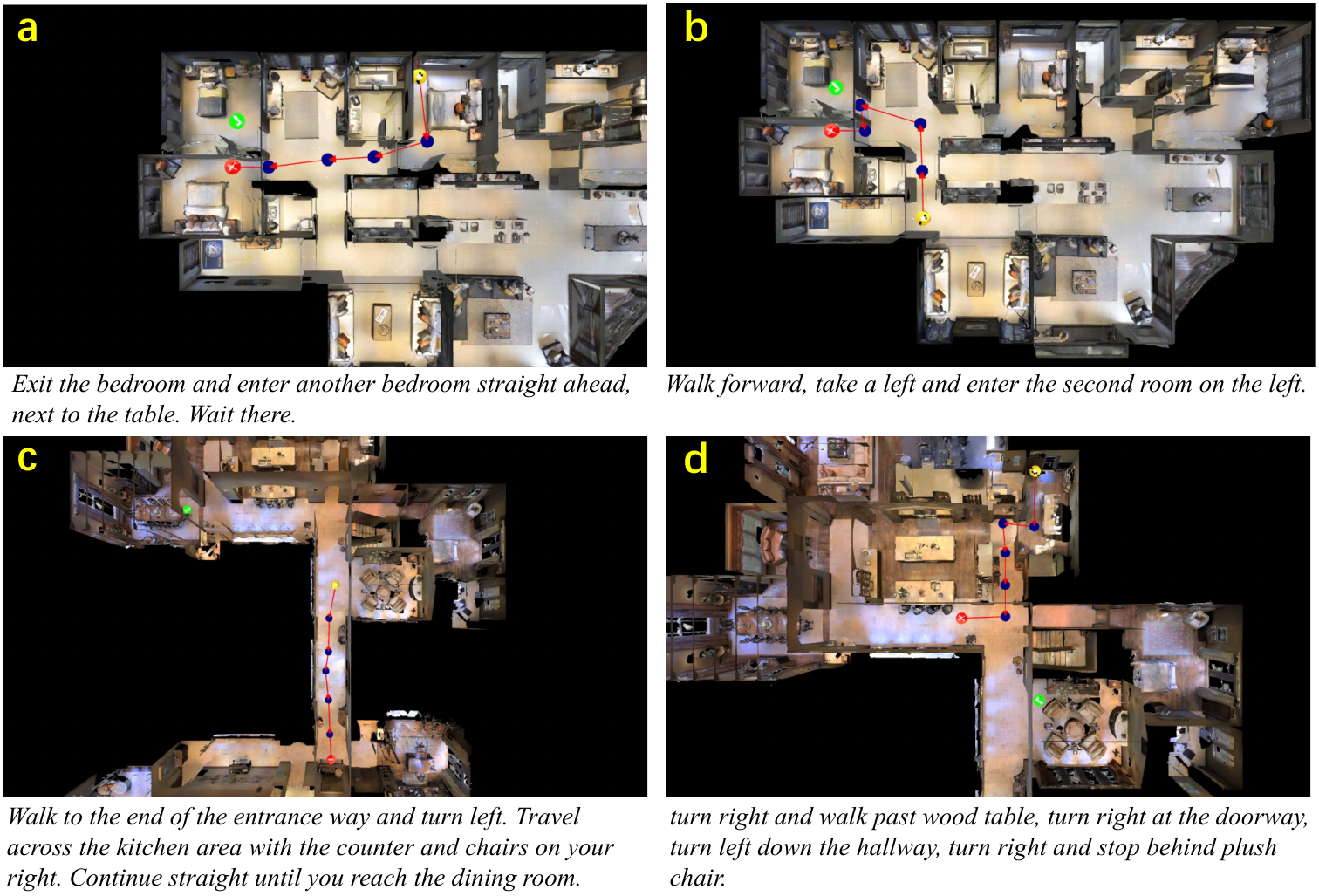}
\vspace{-3mm}
\caption{
Failure cases of `ambiguity' in val unseen splits of R2R. Yellow and green circles denote the start and target locations, respectively, and the red circles represent incorrect endpoints. 
}\label{fig:ambiguity}
\end{figure*}

We visualize some failure cases in Fig.~\ref{fig:early} and Fig.~\ref{fig:ambiguity}. We conclude the failure reasons as `early lost' and `ambiguity'.

Fig.~\ref{fig:early} presents four `early lost' cases. 
The agent loses the state tracking of navigation due to early mistakes, leading to too much backtracking in cases (a,b,c). 
However, it does not go back to the right path till the end. 
In case (d), the agent does not ``turn left'' after ``into the hallway''. This does not trigger backtracking, but the agent directly ``wait by the kitchen counter'' at the wrong location after seeing a counter.

Fig.~\ref{fig:ambiguity} shows four `ambiguity' cases. 
Some ambiguous instructions may confuse the agent, such as ``enter another bedroom straight ahead'' in case (a) and ``enter the second room on the left'' in case (b). 
In case (c), the agent ``walk to the end of the entrance way'' in the opposite direction. After reaching the hallway end, it has lost state tracking and cannot backtrack. 
In case (d), the agent does not know whether it has finished ``down the hallway'' or not, then makes an early ``turn right'' and stops in advance.

%% file: _main.bbl
\begin{thebibliography}{10}

\bibitem{anderson2018vision}
Peter Anderson, Qi~Wu, Damien Teney, Jake Bruce, Mark Johnson, Niko
  S{\"u}nderhauf, Ian Reid, Stephen Gould, and Anton Van Den~Hengel.
\newblock Vision-and-language navigation: Interpreting visually-grounded
  navigation instructions in real environments.
\newblock In {\em Proceedings of the IEEE conference on computer vision and
  pattern recognition}, pages 3674--3683, 2018.

\bibitem{qi2020reverie}
Yuankai Qi, Qi~Wu, Peter Anderson, Xin Wang, William~Yang Wang, Chunhua Shen,
  and Anton van~den Hengel.
\newblock Reverie: Remote embodied visual referring expression in real indoor
  environments.
\newblock In {\em Proceedings of the IEEE/CVF Conference on Computer Vision and
  Pattern Recognition}, pages 9982--9991, 2020.

\bibitem{ku2020room}
Alexander Ku, Peter Anderson, Roma Patel, Eugene Ie, and Jason Baldridge.
\newblock Room-across-room: Multilingual vision-and-language navigation with
  dense spatiotemporal grounding.
\newblock In {\em Proceedings of the 2020 Conference on Empirical Methods in
  Natural Language Processing (EMNLP)}, pages 4392--4412, 2020.

\bibitem{chen2020uniter}
Yen-Chun Chen, Linjie Li, Licheng Yu, Ahmed El~Kholy, Faisal Ahmed, Zhe Gan,
  Yu~Cheng, and Jingjing Liu.
\newblock Uniter: Universal image-text representation learning.
\newblock In {\em Computer Vision--ECCV 2020: 16th European Conference,
  Glasgow, UK, August 23--28, 2020, Proceedings, Part XXX}, pages 104--120.
  Springer, 2020.

\bibitem{tan2019lxmert}
Hao Tan and Mohit Bansal.
\newblock Lxmert: Learning cross-modality encoder representations from
  transformers.
\newblock In {\em Proceedings of the 2019 Conference on Empirical Methods in
  Natural Language Processing and the 9th International Joint Conference on
  Natural Language Processing (EMNLP-IJCNLP)}, pages 5100--5111, 2019.

\bibitem{chen2023vlp}
Fei-Long Chen, Du-Zhen Zhang, Ming-Lun Han, Xiu-Yi Chen, Jing Shi, Shuang Xu,
  and Bo~Xu.
\newblock Vlp: A survey on vision-language pre-training.
\newblock {\em Machine Intelligence Research}, 20(1):38--56, 2023.

\bibitem{ji2023masked}
Ge-Peng Ji, Mingchen Zhuge, Dehong Gao, Deng-Ping Fan, Christos Sakaridis, and
  Luc~Van Gool.
\newblock Masked vision-language transformer in fashion.
\newblock {\em Machine Intelligence Research}, 20(3):421--434, 2023.

\bibitem{gu2023eva2}
Yuxian Gu, Jiaxin Wen, Hao Sun, Yi~Song, Pei Ke, Chujie Zheng, Zheng Zhang,
  Jianzhu Yao, Lei Liu, Xiaoyan Zhu, et~al.
\newblock Eva2. 0: Investigating open-domain chinese dialogue systems with
  large-scale pre-training.
\newblock {\em Machine Intelligence Research}, 20(2):207--219, 2023.

\bibitem{wang2023large}
Xiao Wang, Guangyao Chen, Guangwu Qian, Pengcheng Gao, Xiao-Yong Wei, Yaowei
  Wang, Yonghong Tian, and Wen Gao.
\newblock Large-scale multi-modal pre-trained models: A comprehensive survey.
\newblock {\em Machine Intelligence Research}, pages 1--36, 2023.

\bibitem{hao2020towards}
Weituo Hao, Chunyuan Li, Xiujun Li, Lawrence Carin, and Jianfeng Gao.
\newblock Towards learning a generic agent for vision-and-language navigation
  via pre-training.
\newblock In {\em Proceedings of the IEEE/CVF Conference on Computer Vision and
  Pattern Recognition}, pages 13137--13146, 2020.

\bibitem{majumdar2020improving}
Arjun Majumdar, Ayush Shrivastava, Stefan Lee, Peter Anderson, Devi Parikh, and
  Dhruv Batra.
\newblock Improving vision-and-language navigation with image-text pairs from
  the web.
\newblock In {\em European Conference on Computer Vision}, pages 259--274.
  Springer, 2020.

\bibitem{guhur2021airbert}
Pierre-Louis Guhur, Makarand Tapaswi, Shizhe Chen, Ivan Laptev, and Cordelia
  Schmid.
\newblock Airbert: In-domain pretraining for vision-and-language navigation.
\newblock In {\em Proceedings of the IEEE/CVF International Conference on
  Computer Vision}, pages 1634--1643, 2021.

\bibitem{qi2021road}
Yuankai Qi, Zizheng Pan, Yicong Hong, Ming-Hsuan Yang, Anton van~den Hengel,
  and Qi~Wu.
\newblock The road to know-where: An object-and-room informed sequential bert
  for indoor vision-language navigation.
\newblock In {\em Proceedings of the IEEE/CVF International Conference on
  Computer Vision}, pages 1655--1664, 2021.

\bibitem{qiao2022hop}
Yanyuan Qiao, Yuankai Qi, Yicong Hong, Zheng Yu, Peng Wang, and Qi~Wu.
\newblock Hop: History-and-order aware pre-training for vision-and-language
  navigation.
\newblock In {\em Proceedings of the IEEE/CVF Conference on Computer Vision and
  Pattern Recognition}, pages 15418--15427, 2022.

\bibitem{chenweakly}
Peihao Chen, Dongyu Ji, Kunyang Lin, Runhao Zeng, Thomas~H Li, Mingkui Tan, and
  Chuang Gan.
\newblock Weakly-supervised multi-granularity map learning for
  vision-and-language navigation.
\newblock In {\em Advances in Neural Information Processing Systems}.

\bibitem{chaplot2020object}
Devendra~Singh Chaplot, Dhiraj~Prakashchand Gandhi, Abhinav Gupta, and Russ~R
  Salakhutdinov.
\newblock Object goal navigation using goal-oriented semantic exploration.
\newblock {\em Advances in Neural Information Processing Systems},
  33:4247--4258, 2020.

\bibitem{chaplot2020neural}
Devendra~Singh Chaplot, Ruslan Salakhutdinov, Abhinav Gupta, and Saurabh Gupta.
\newblock Neural topological slam for visual navigation.
\newblock In {\em Proceedings of the IEEE/CVF Conference on Computer Vision and
  Pattern Recognition}, pages 12875--12884, 2020.

\bibitem{henriques2018mapnet}
Joao~F Henriques and Andrea Vedaldi.
\newblock Mapnet: An allocentric spatial memory for mapping environments.
\newblock In {\em proceedings of the IEEE Conference on Computer Vision and
  Pattern Recognition}, pages 8476--8484, 2018.

\bibitem{wang2021structured}
Hanqing Wang, Wenguan Wang, Wei Liang, Caiming Xiong, and Jianbing Shen.
\newblock Structured scene memory for vision-language navigation.
\newblock In {\em Proceedings of the IEEE/CVF Conference on Computer Vision and
  Pattern Recognition}, pages 8455--8464, 2021.

\bibitem{konolige2011navigation}
Kurt Konolige, Eitan Marder-Eppstein, and Bhaskara Marthi.
\newblock Navigation in hybrid metric-topological maps.
\newblock In {\em 2011 IEEE International Conference on Robotics and
  Automation}, pages 3041--3047. IEEE, 2011.

\bibitem{georgakis2022cross}
Georgios Georgakis, Karl Schmeckpeper, Karan Wanchoo, Soham Dan, Eleni
  Miltsakaki, Dan Roth, and Kostas Daniilidis.
\newblock Cross-modal map learning for vision and language navigation.
\newblock In {\em Proceedings of the IEEE/CVF Conference on Computer Vision and
  Pattern Recognition}, pages 15460--15470, 2022.

\bibitem{chen2021history}
Shizhe Chen, Pierre-Louis Guhur, Cordelia Schmid, and Ivan Laptev.
\newblock History aware multimodal transformer for vision-and-language
  navigation.
\newblock {\em Advances in Neural Information Processing Systems},
  34:5834--5847, 2021.

\bibitem{deng2020evolving}
Zhiwei Deng, Karthik Narasimhan, and Olga Russakovsky.
\newblock Evolving graphical planner: Contextual global planning for
  vision-and-language navigation.
\newblock {\em Advances in Neural Information Processing Systems},
  33:20660--20672, 2020.

\bibitem{chen2022think}
Shizhe Chen, Pierre-Louis Guhur, Makarand Tapaswi, Cordelia Schmid, and Ivan
  Laptev.
\newblock Think global, act local: Dual-scale graph transformer for
  vision-and-language navigation.
\newblock In {\em Proceedings of the IEEE/CVF Conference on Computer Vision and
  Pattern Recognition}, pages 16537--16547, 2022.

\bibitem{blanco2008toward}
Jose-Luis Blanco, Juan-Antonio Fern{\'a}ndez-Madrigal, and Javier Gonzalez.
\newblock Toward a unified bayesian approach to hybrid metric--topological
  slam.
\newblock {\em IEEE Transactions on Robotics}, 24(2):259--270, 2008.

\bibitem{kenton2019bert}
Jacob Devlin Ming-Wei~Chang Kenton and Lee~Kristina Toutanova.
\newblock Bert: Pre-training of deep bidirectional transformers for language
  understanding.
\newblock In {\em Proceedings of NAACL-HLT}, pages 4171--4186, 2019.

\bibitem{krantz2020beyond}
Jacob Krantz, Erik Wijmans, Arjun Majumdar, Dhruv Batra, and Stefan Lee.
\newblock Beyond the nav-graph: Vision-and-language navigation in continuous
  environments.
\newblock In {\em European Conference on Computer Vision}, pages 104--120.
  Springer, 2020.

\bibitem{he2021landmark}
Keji He, Yan Huang, Qi~Wu, Jianhua Yang, Dong An, Shuanglin Sima, and Liang
  Wang.
\newblock Landmark-rxr: Solving vision-and-language navigation with
  fine-grained alignment supervision.
\newblock {\em Advances in Neural Information Processing Systems}, 34:652--663,
  2021.

\bibitem{gu2022vision}
Jing Gu, Eliana Stefani, Qi~Wu, Jesse Thomason, and Xin Wang.
\newblock Vision-and-language navigation: A survey of tasks, methods, and
  future directions.
\newblock In {\em Proceedings of the 60th Annual Meeting of the Association for
  Computational Linguistics (Volume 1: Long Papers)}, pages 7606--7623, 2022.

\bibitem{wang2022towards}
Hanqing Wang, Wei Liang, Luc~V Gool, and Wenguan Wang.
\newblock Towards versatile embodied navigation.
\newblock {\em Advances in Neural Information Processing Systems},
  35:36858--36874, 2022.

\bibitem{Zhu2022diagnosing}
Wanrong Zhu, Yuankai Qi, P.~Narayana, Kazoo Sone, Sugato Basu, Xin~Eric Wang,
  Qi~Wu, Miguel~P. Eckstein, and William~Yang Wang.
\newblock Diagnosing vision-and-language navigation: What really matters.
\newblock In {\em NAACL}, 2022.

\bibitem{fried2018speaker}
Daniel Fried, Ronghang Hu, Volkan Cirik, Anna Rohrbach, Jacob Andreas,
  Louis-Philippe Morency, Taylor Berg-Kirkpatrick, Kate Saenko, Dan Klein, and
  Trevor Darrell.
\newblock Speaker-follower models for vision-and-language navigation.
\newblock {\em Advances in Neural Information Processing Systems}, 31, 2018.

\bibitem{ma2019self}
Chih-Yao Ma, Jiasen Lu, Zuxuan Wu, Ghassan AlRegib, Zsolt Kira, Richard Socher,
  and Caiming Xiong.
\newblock Self-monitoring navigation agent via auxiliary progress estimation.
\newblock {\em arXiv preprint arXiv:1901.03035}, 2019.

\bibitem{qi2020object}
Yuankai Qi, Zizheng Pan, Shengping Zhang, Anton van~den Hengel, and Qi~Wu.
\newblock Object-and-action aware model for visual language navigation.
\newblock In {\em European Conference on Computer Vision}, pages 303--317.
  Springer, 2020.

\bibitem{hong2020language}
Yicong Hong, Cristian Rodriguez, Yuankai Qi, Qi~Wu, and Stephen Gould.
\newblock Language and visual entity relationship graph for agent navigation.
\newblock {\em Advances in Neural Information Processing Systems},
  33:7685--7696, 2020.

\bibitem{an2021neighbor}
Dong An, Yuankai Qi, Yan Huang, Qi~Wu, Liang Wang, and Tieniu Tan.
\newblock Neighbor-view enhanced model for vision and language navigation.
\newblock In {\em Proceedings of the 29th ACM International Conference on
  Multimedia}, pages 5101--5109, 2021.

\bibitem{wang2018look}
Xin Wang, Wenhan Xiong, Hongmin Wang, and William~Yang Wang.
\newblock Look before you leap: Bridging model-free and model-based
  reinforcement learning for planned-ahead vision-and-language navigation.
\newblock In {\em Proceedings of the European Conference on Computer Vision
  (ECCV)}, pages 37--53, 2018.

\bibitem{wang2019reinforced}
Xin Wang, Qiuyuan Huang, Asli Celikyilmaz, Jianfeng Gao, Dinghan Shen,
  Yuan-Fang Wang, William~Yang Wang, and Lei Zhang.
\newblock Reinforced cross-modal matching and self-supervised imitation
  learning for vision-language navigation.
\newblock In {\em Proceedings of the IEEE/CVF Conference on Computer Vision and
  Pattern Recognition}, pages 6629--6638, 2019.

\bibitem{tan2019learning}
Hao Tan, Licheng Yu, and Mohit Bansal.
\newblock Learning to navigate unseen environments: Back translation with
  environmental dropout.
\newblock In {\em Proceedings of NAACL-HLT}, pages 2610--2621, 2019.

\bibitem{liu2021vision}
Chong Liu, Fengda Zhu, Xiaojun Chang, Xiaodan Liang, Zongyuan Ge, and Yi-Dong
  Shen.
\newblock Vision-language navigation with random environmental mixup.
\newblock In {\em Proceedings of the IEEE/CVF International Conference on
  Computer Vision}, pages 1644--1654, 2021.

\bibitem{koh2021pathdreamer}
Jing~Yu Koh, Honglak Lee, Yinfei Yang, Jason Baldridge, and Peter Anderson.
\newblock Pathdreamer: A world model for indoor navigation.
\newblock In {\em Proceedings of the IEEE/CVF International Conference on
  Computer Vision}, pages 14738--14748, 2021.

\bibitem{li2022envedit}
Jialu Li, Hao Tan, and Mohit Bansal.
\newblock Envedit: Environment editing for vision-and-language navigation.
\newblock In {\em Proceedings of the IEEE/CVF Conference on Computer Vision and
  Pattern Recognition}, pages 15407--15417, 2022.

\bibitem{li2023improving}
Jialu Li and Mohit Bansal.
\newblock Improving vision-and-language navigation by generating future-view
  image semantics.
\newblock In {\em Proceedings of the IEEE/CVF Conference on Computer Vision and
  Pattern Recognition}, pages 10803--10812, 2023.

\bibitem{wang2023scale}
Zun Wang, Jialu Li, Yicong Hong, Yi~Wang, Qi~Wu, Mohit Bansal, Stephen Gould,
  Hao Tan, and Yu~Qiao.
\newblock Scaling data generation in vision-and-language navigation.
\newblock In {\em Proceedings of the IEEE/CVF International Conference on
  Computer Vision}, 2023.

\bibitem{wang2023lana}
Xiaohan Wang, Wenguan Wang, Jiayi Shao, and Yi~Yang.
\newblock Lana: A language-capable navigator for instruction following and
  generation.
\newblock In {\em Proceedings of the IEEE/CVF Conference on Computer Vision and
  Pattern Recognition}, pages 19048--19058, 2023.

\bibitem{wang2022counterfactual}
Hanqing Wang, Wei Liang, Jianbing Shen, Luc Van~Gool, and Wenguan Wang.
\newblock Counterfactual cycle-consistent learning for instruction following
  and generation in vision-language navigation.
\newblock In {\em Proceedings of the IEEE/CVF Conference on Computer Vision and
  Pattern Recognition}, pages 15471--15481, 2022.

\bibitem{chen2022learning}
Shizhe Chen, Pierre-Louis Guhur, Makarand Tapaswi, Cordelia Schmid, and Ivan
  Laptev.
\newblock Learning from unlabeled 3d environments for vision-and-language
  navigation.
\newblock In {\em Computer Vision--ECCV 2022: 17th European Conference, Tel
  Aviv, Israel, October 23--27, 2022, Proceedings, Part XXXIX}, pages 638--655.
  Springer, 2022.

\bibitem{hong2021vln}
Yicong Hong, Qi~Wu, Yuankai Qi, Cristian Rodriguez-Opazo, and Stephen Gould.
\newblock Vln bert: A recurrent vision-and-language bert for navigation.
\newblock In {\em Proceedings of the IEEE/CVF Conference on Computer Vision and
  Pattern Recognition}, pages 1643--1653, 2021.

\bibitem{zhao2022target}
Yusheng Zhao, Jinyu Chen, Chen Gao, Wenguan Wang, Lirong Yang, Haibing Ren,
  Huaxia Xia, and Si~Liu.
\newblock Target-driven structured transformer planner for vision-language
  navigation.
\newblock In {\em Proceedings of the 30th ACM International Conference on
  Multimedia}, pages 4194--4203, 2022.

\bibitem{an2023etpnav}
Dong An, Hanqing Wang, Wenguan Wang, Zun Wang, Yan Huang, Keji He, and Liang
  Wang.
\newblock Etpnav: Evolving topological planning for vision-language navigation
  in continuous environments.
\newblock {\em arXiv preprint arXiv:2304.03047}, 2023.

\bibitem{radford2021learning}
Alec Radford, Jong~Wook Kim, Chris Hallacy, Aditya Ramesh, Gabriel Goh,
  Sandhini Agarwal, Girish Sastry, Amanda Askell, Pamela Mishkin, Jack Clark,
  et~al.
\newblock Learning transferable visual models from natural language
  supervision.
\newblock In {\em International Conference on Machine Learning}, pages
  8748--8763. PMLR, 2021.

\bibitem{lu2019vilbert}
Jiasen Lu, Dhruv Batra, Devi Parikh, and Stefan Lee.
\newblock Vilbert: Pretraining task-agnostic visiolinguistic representations
  for vision-and-language tasks.
\newblock {\em Advances in neural information processing systems}, 32, 2019.

\bibitem{anderson2018bottom}
Peter Anderson, Xiaodong He, Chris Buehler, Damien Teney, Mark Johnson, Stephen
  Gould, and Lei Zhang.
\newblock Bottom-up and top-down attention for image captioning and visual
  question answering.
\newblock In {\em Proceedings of the IEEE conference on computer vision and
  pattern recognition}, pages 6077--6086, 2018.

\bibitem{ren2015faster}
Shaoqing Ren, Kaiming He, Ross Girshick, and Jian Sun.
\newblock Faster r-cnn: Towards real-time object detection with region proposal
  networks.
\newblock {\em Advances in neural information processing systems}, 28, 2015.

\bibitem{jiang2020defense}
Huaizu Jiang, Ishan Misra, Marcus Rohrbach, Erik Learned-Miller, and Xinlei
  Chen.
\newblock In defense of grid features for visual question answering.
\newblock In {\em Proceedings of the IEEE/CVF Conference on Computer Vision and
  Pattern Recognition}, pages 10267--10276, 2020.

\bibitem{huang2021seeing}
Zhicheng Huang, Zhaoyang Zeng, Yupan Huang, Bei Liu, Dongmei Fu, and Jianlong
  Fu.
\newblock Seeing out of the box: End-to-end pre-training for vision-language
  representation learning.
\newblock In {\em Proceedings of the IEEE/CVF Conference on Computer Vision and
  Pattern Recognition}, pages 12976--12985, 2021.

\bibitem{fuentes2015visual}
Jorge Fuentes-Pacheco, Jos{\'e} Ruiz-Ascencio, and Juan~Manuel
  Rend{\'o}n-Mancha.
\newblock Visual simultaneous localization and mapping: a survey.
\newblock {\em Artificial intelligence review}, 43(1):55--81, 2015.

\bibitem{narasimhan2020seeing}
Medhini Narasimhan, Erik Wijmans, Xinlei Chen, Trevor Darrell, Dhruv Batra,
  Devi Parikh, and Amanpreet Singh.
\newblock Seeing the un-scene: Learning amodal semantic maps for room
  navigation.
\newblock In {\em Computer Vision--ECCV 2020: 16th European Conference,
  Glasgow, UK, August 23--28, 2020, Proceedings, Part XVIII 16}, pages
  513--529. Springer, 2020.

\bibitem{irshad2021sasra}
Muhammad~Zubair Irshad, Niluthpol~Chowdhury Mithun, Zachary Seymour, Han-Pang
  Chiu, Supun Samarasekera, and Rakesh Kumar.
\newblock Sasra: Semantically-aware spatio-temporal reasoning agent for
  vision-and-language navigation in continuous environments.
\newblock {\em arXiv preprint arXiv:2108.11945}, 2021.

\bibitem{chen2021topological}
Kevin Chen, Junshen~K Chen, Jo~Chuang, Marynel V{\'a}zquez, and Silvio
  Savarese.
\newblock Topological planning with transformers for vision-and-language
  navigation.
\newblock In {\em Proceedings of the IEEE/CVF Conference on Computer Vision and
  Pattern Recognition}, pages 11276--11286, 2021.

\bibitem{kwon2021visual}
Obin Kwon, Nuri Kim, Yunho Choi, Hwiyeon Yoo, Jeongho Park, and Songhwai Oh.
\newblock Visual graph memory with unsupervised representation for visual
  navigation.
\newblock In {\em Proceedings of the IEEE/CVF International Conference on
  Computer Vision}, pages 15890--15899, 2021.

\bibitem{anderson2019chasing}
Peter Anderson, Ayush Shrivastava, Devi Parikh, Dhruv Batra, and Stefan Lee.
\newblock Chasing ghosts: Instruction following as bayesian state tracking.
\newblock {\em Advances in neural information processing systems}, 32, 2019.

\bibitem{dosovitskiy2020image}
Alexey Dosovitskiy, Lucas Beyer, Alexander Kolesnikov, Dirk Weissenborn,
  Xiaohua Zhai, Thomas Unterthiner, Mostafa Dehghani, Matthias Minderer, Georg
  Heigold, Sylvain Gelly, et~al.
\newblock An image is worth 16x16 words: Transformers for image recognition at
  scale.
\newblock In {\em International Conference on Learning Representations}, 2020.

\bibitem{chang2017matterport3d}
Angel Chang, Angela Dai, Thomas Funkhouser, Maciej Halber, Matthias Niebner,
  Manolis Savva, Shuran Song, Andy Zeng, and Yinda Zhang.
\newblock Matterport3d: Learning from rgb-d data in indoor environments.
\newblock In {\em 2017 International Conference on 3D Vision (3DV)}, pages
  667--676. IEEE, 2017.

\bibitem{anderson2018evaluation}
Peter Anderson, Angel Chang, Devendra~Singh Chaplot, Alexey Dosovitskiy,
  Saurabh Gupta, Vladlen Koltun, Jana Kosecka, Jitendra Malik, Roozbeh
  Mottaghi, Manolis Savva, et~al.
\newblock On evaluation of embodied navigation agents.
\newblock {\em arXiv preprint arXiv:1807.06757}, 2018.

\bibitem{ilharco2019general}
Gabriel Ilharco, Vihan Jain, Alexander Ku, Eugene Ie, and Jason Baldridge.
\newblock General evaluation for instruction conditioned navigation using
  dynamic time warping.
\newblock {\em arXiv preprint arXiv:1907.05446}, 2019.

\bibitem{shen2021much}
Sheng Shen, Liunian~Harold Li, Hao Tan, Mohit Bansal, Anna Rohrbach, Kai-Wei
  Chang, Zhewei Yao, and Kurt Keutzer.
\newblock How much can clip benefit vision-and-language tasks?
\newblock In {\em International Conference on Learning Representations}, 2021.

\bibitem{liu2019roberta}
Yinhan Liu, Myle Ott, Naman Goyal, Jingfei Du, Mandar Joshi, Danqi Chen, Omer
  Levy, Mike Lewis, Luke Zettlemoyer, and Veselin Stoyanov.
\newblock Roberta: A robustly optimized bert pretraining approach.
\newblock {\em arXiv preprint arXiv:1907.11692}, 2019.

\bibitem{wang2022less}
Su~Wang, Ceslee Montgomery, Jordi Orbay, Vighnesh Birodkar, Aleksandra Faust,
  Izzeddin Gur, Natasha Jaques, Austin Waters, Jason Baldridge, and Peter
  Anderson.
\newblock Less is more: Generating grounded navigation instructions from
  landmarks.
\newblock In {\em Proceedings of the IEEE/CVF Conference on Computer Vision and
  Pattern Recognition}, pages 15428--15438, 2022.

\bibitem{hong2022bridging}
Yicong Hong, Zun Wang, Qi~Wu, and Stephen Gould.
\newblock Bridging the gap between learning in discrete and continuous
  environments for vision-and-language navigation.
\newblock In {\em Proceedings of the IEEE/CVF Conference on Computer Vision and
  Pattern Recognition}, pages 15439--15449, 2022.

\bibitem{zhu2020vision}
Fengda Zhu, Yi~Zhu, Xiaojun Chang, and Xiaodan Liang.
\newblock Vision-language navigation with self-supervised auxiliary reasoning
  tasks.
\newblock In {\em Proceedings of the IEEE/CVF Conference on Computer Vision and
  Pattern Recognition}, pages 10012--10022, 2020.

\bibitem{qiao2023hop+}
Yanyuan Qiao, Yuankai Qi, Yicong Hong, Zheng Yu, Peng Wang, and Qi~Wu.
\newblock Hop+: History-enhanced and order-aware pre-training for
  vision-and-language navigation.
\newblock {\em IEEE Transactions on Pattern Analysis and Machine Intelligence},
  2023.

\bibitem{krantz2021waypoint}
Jacob Krantz, Aaron Gokaslan, Dhruv Batra, Stefan Lee, and Oleksandr Maksymets.
\newblock Waypoint models for instruction-guided navigation in continuous
  environments.
\newblock In {\em Proceedings of the IEEE/CVF International Conference on
  Computer Vision}, pages 15162--15171, 2021.

\bibitem{krantz2022sim}
Jacob Krantz and Stefan Lee.
\newblock Sim-2-sim transfer for vision-and-language navigation in continuous
  environments.
\newblock In {\em Computer Vision--ECCV 2022: 17th European Conference, Tel
  Aviv, Israel, October 23--27, 2022, Proceedings, Part XXXIX}, pages 588--603.
  Springer, 2022.

\bibitem{an20221st}
Dong An, Zun Wang, Yangguang Li, Yi~Wang, Yicong Hong, Yan Huang, Liang Wang,
  and Jing Shao.
\newblock 1st place solutions for rxr-habitat vision-and-language navigation
  competition (cvpr 2022).
\newblock {\em arXiv preprint arXiv:2206.11610}, 2022.

\bibitem{li2022clear}
Jialu Li, Hao Tan, and Mohit Bansal.
\newblock Clear: Improving vision-language navigation with cross-lingual,
  environment-agnostic representations.
\newblock In {\em Findings of the Association for Computational Linguistics:
  NAACL 2022}, pages 633--649, 2022.

\bibitem{lin2021scene}
Xiangru Lin, Guanbin Li, and Yizhou Yu.
\newblock Scene-intuitive agent for remote embodied visual grounding.
\newblock In {\em Proceedings of the IEEE/CVF Conference on Computer Vision and
  Pattern Recognition}, pages 7036--7045, 2021.

\bibitem{wijmansdd}
Erik Wijmans, Abhishek Kadian, Ari Morcos, Stefan Lee, Irfan Essa, Devi Parikh,
  Manolis Savva, and Dhruv Batra.
\newblock Dd-ppo: Learning near-perfect pointgoal navigators from 2.5 billion
  frames.
\newblock In {\em International Conference on Learning Representations}.

\bibitem{jiang2018rednet}
Jindong Jiang, Lunan Zheng, Fei Luo, and Zhijun Zhang.
\newblock Rednet: Residual encoder-decoder network for indoor rgb-d semantic
  segmentation.
\newblock {\em arXiv preprint arXiv:1806.01054}, 2018.

\bibitem{koh2022simple}
Jing~Yu Koh, Harsh Agrawal, Dhruv Batra, Richard Tucker, Austin Waters, Honglak
  Lee, Yinfei Yang, Jason Baldridge, and Peter Anderson.
\newblock Simple and effective synthesis of indoor 3d scenes.

\bibitem{deng2009imagenet}
Jia Deng, Wei Dong, Richard Socher, Li-Jia Li, Kai Li, and Li~Fei-Fei.
\newblock Imagenet: A large-scale hierarchical image database.
\newblock In {\em 2009 IEEE conference on computer vision and pattern
  recognition}, pages 248--255. Ieee, 2009.

\bibitem{savva2019habitat}
Manolis Savva, Abhishek Kadian, Oleksandr Maksymets, Yili Zhao, Erik Wijmans,
  Bhavana Jain, Julian Straub, Jia Liu, Vladlen Koltun, Jitendra Malik, et~al.
\newblock Habitat: A platform for embodied ai research.
\newblock In {\em Proceedings of the IEEE/CVF International Conference on
  Computer Vision}, pages 9339--9347, 2019.

\bibitem{wang2020active}
Hanqing Wang, Wenguan Wang, Tianmin Shu, Wei Liang, and Jianbing Shen.
\newblock Active visual information gathering for vision-language navigation.
\newblock In {\em European Conference on Computer Vision}, pages 307--322.
  Springer, 2020.

\bibitem{chen2022reinforced}
Jinyu Chen, Chen Gao, Erli Meng, Qiong Zhang, and Si~Liu.
\newblock Reinforced structured state-evolution for vision-language navigation.
\newblock In {\em Proceedings of the IEEE/CVF Conference on Computer Vision and
  Pattern Recognition}, pages 15450--15459, 2022.

\end{thebibliography}
